\documentclass{article}

    \usepackage[preprint,nonatbib]{neurips_2020}

\usepackage[utf8]{inputenc} 
\usepackage[T1]{fontenc}    
\usepackage{hyperref}       
\usepackage{url}            
\usepackage{booktabs}       
\usepackage{amsfonts}       
\usepackage{nicefrac}       
\usepackage{microtype}      
\usepackage{graphicx,caption,subcaption}
\usepackage{amsmath}
\usepackage{wrapfig}

\title{One-vs-Rest Network-based Deep Probability Model for Open Set Recognition}

\author{%
  Jaeyeon~Jang \qquad Chang Ouk Kim \\
  Department of Industrial Engineering\\
  Yonsei University\\
  \texttt{\{jjy009,kimco\}@yonsei.ac.kr} \\
}

\begin{document}

\maketitle

\begin{abstract}
Unknown examples that are unseen during training often appear in real-world computer vision tasks, and an intelligent self-learning system should be able to differentiate between known and unknown examples. Open set recognition, which addresses this problem, has been studied for approximately a decade. However, conventional open set recognition methods based on deep neural networks (DNNs) lack a foundation for post recognition score analysis. In this paper, we propose a DNN structure in which multiple one-vs-rest sigmoid networks follow a convolutional neural network feature extractor. A one-vs-rest network, which is composed of rectified linear unit activation functions for the hidden layers and a single sigmoid target class output node, can maximize the ability to learn information from nonmatch examples. Furthermore, the network yields a sophisticated nonlinear features-to-output mapping that is explainable in the feature space. By introducing extreme value theory-based calibration techniques, the nonlinear and explainable mapping provides a well-grounded class membership probability models. Our experiments show that one-vs-rest networks can provide more informative hidden representations for unknown examples than the commonly used SoftMax layer. In addition, the proposed probability model outperformed the state-of-the art methods in open set classification scenarios.
\end{abstract}

\section{Introduction}

Recent advancements in deep learning have greatly improved the performance of visual recognition systems \cite{He2016,Parkhi2015,Simonyan2015,Krizhevsky2012}. In particular, the latest deep convolutional neural network (CNN) models have surpassed human-level performance in terms of classification error rates \cite{He2015}. However, the vast majority of recognition systems are designed under closed-world assumptions in which all categories are known a priori. Although this assumption holds in many applications, the need to detect objects unknown at training time, which is called open set recognition, has recently been highlighted \cite{Zhang2017,Spigler2019,Rudd2018,Scheirer2014}. For example, self-driving cars must be able to recognize new objects to respond as they emerge. Furthermore, the ability to differentiate between known and unknown has been considered a key element of intelligent self-learning systems \cite{Boult2019}.

Conventional closed set classifiers have been developed for modeling the optimal posterior probability $P(y\mid x;y\in \Upsilon )$, where $x$ is an input sample and $\Upsilon$ is the set of known class indices. The classifiers assign the class  $y$ that maximizes $P(y\mid x)$ given an input sample $x$. In this scenario, a sample of an unknown class will be misclassified as one of the known classes. Thus, the objective of an open set recognition system is to provide a criterion to accept a sample as the highest confidence label or reject a sample as unknown. Additionally, to construct a well-performing open set recognition system, the system should be able to minimize open space risk, which is the relative measure of positively labeled space where target known samples are unlikely to exist in the entire positively labeled space \cite{Scheirer2014,Scheirer2013}.

In general, open set recognition is composed of producing recognition scores for known classes and a subsequent post score analysis. Extreme value theory (EVT)-based calibration methods have been widely adopted for post recognition score analysis \cite{Bendale2016,Oza2019,Ge2017}. EVT is a branch of statistical theory related to the distribution of extreme examples. The theory has been applied to calibrate recognition scores based on the distributions of match extremes and nonmatch extremes to provide balanced accept/reject criteria for all known classes \cite{Scheirer2011}. For example, Scheirer et al. \cite{Scheirer2014} applied the EVT-based Weibull calibration method to decision scores of a one-vs-rest support vector machine (SVM) for multiclass open set recognition. They also used one-class SVM decision scores to support the decision scores of one-vs-rest SVM, providing theoretical background that one-class SVM can manage open space risk. Bendale et al. \cite{Bendale2016} applied an EVT-based calibration strategy to decision scores calculated from the penultimate layer of a CNN. After applying a calibration method, they could obtain a probability score called OpenMax that takes into account open space risk in the feature space. Additionally, the generalized Pareto distribution, an extreme value distribution, was used to estimate the tail distribution for errors of reconstruction models in \cite{Zhang2017} and \cite{Oza2019}.

Open set recognition is seen as a well-established area, but only for methods that apply shallow machine learning models such as SVM. For deep neural network (DNN) models, there is a lack of theoretical foundation in adopting EVT-based post recognition score analysis for open set recognition \cite{Boult2019}. Generally, EVT-based calibration produces class membership probabilities based on the recognition score. For the probability to be reliable, the DNN producing the score must satisfy the following conditions. First, the sophisticated nonlinear features-to-score mapping should be modeled to yield a sophisticated features-to-probability mapping. Second, increases and decreases in the score should be explained explicitly based on the distance from positive sample points in the feature space. Thus, it should be guaranteed that a probability model can be constructed on the feature space. Finally, and most importantly, the DNN should be able to minimize and thus manage the open space risk. In addition to these conditions, for an open set recognition system to be introduced into an intelligent self-learning system, the DNN should be able to retrieve useful information from unknown examples. This is partially possible by maximizing the ability of learning information from nonmatch classes. Thus, in this paper, we propose a DNN structure that is designed not only to improve the recognition performance but also to extract more information from nonmatch samples during training, providing a well-grounded theoretical foundation for post analysis of the scores produced by the proposed model.

We propose a model that combines a CNN feature extractor and the following one-vs-rest sigmoid networks. The proposed one-vs-rest sigmoid networks, which use rectified linear unit (ReLU) activation for hidden layers, consist of a simple feedforward neural network that produces a sigmoid decision score for each target class, as shown in Fig. \ref{structure} in the supplementary section \ref{spplementary}. In terms of structure, the one-vs-rest sigmoid networks share the same hidden representation. The proposed model can learn more features from nonmatch examples compared to general CNN classifiers that use a SoftMax output layer because each one-vs-rest network learns equally from match and nonmatch samples because of the sigmoid activation. However, the normalization term of SoftMax decreases the ability to learn from nonmatch samples \cite{Oland2017}. Additionally, one-vs-rest sigmoid networks can provide a sophisticated class membership probability mapping from the final hidden feature space by adding nonlinearity. Furthermore, the networks yield a simple monotonically increasing probability function of the distance from the positive sample in each subspace divided by ReLU activation \cite{He2018}. On the basis of this model characteristic, we can derive a compact abating probability (CAP) model in the feature space, where the CAP is a probability that abates as the data point moves from known data to open space. Finally, by incorporating EVT-based calibration into the CAP model, we can minimize open space risk, yielding deep open probability model.

We compared the proposed method with state-of-the-art CNN-based open set recognition methods on three types of real-world image datasets. The experiments showed that a CNN with one-vs-rest sigmoid networks can provide more informative hidden representations for unknown examples than CNN with a SoftMax layer. This result was verified by statistical comparison of the difference in distributional information between known and unknown class examples. Additionally, experiments to measure the classification performance revealed that the proposed method outperformed the state-of-the art approaches in terms of the F-measure.

The remainder of this paper is organized as follows. Section \ref{related} presents a brief review of related works. We provide a theoretical foundation for the deep probability model in Section \ref{model}, in which we also propose post recognition score analysis based on one-vs-rest networks. In Section \ref{experiments}, we evaluate the proposed open set recognition method in terms of various aspects. Finally, Section \ref{conclusion} discusses the results of this paper and future research directions.

\section{Related Work} \label{related}

EVT addresses the characteristics of statistical extremes \cite{Scheirer2010}. The theory makes it possible to model open set recognition systems by providing a theoretical foundation for post recognition score analysis \cite{Scheirer2014,Bendale2016,Scheirer2012}. Generally, for a given recognition model, the scores of the most relevant and important data for prediction lie in the tail \cite{Scheirer2011}. In addition, the threshold to accept/reject usually lies in the overlap region of extremes of match and nonmatch scores \cite{Shi2008}. Thus, we must transform the decision scores provided by a recognition system into probability scores based on the distributional information of the tails: a probability-based rejection/acceptance threshold should be set. EVT opened a new path for open set recognition by providing a well-grounded theorem that the tails of any score distribution must always follow an EVT distribution \cite{Kotz2001}. Since the distribution of extreme scores can be modeled and a class membership probability can be estimated from the distribution model, EVT-based calibration methods have been widely adopted in the field of open set recognition \cite{Oza2019,Perera2017,Gibert2015}.

Open set recognition methods can be divided into two categories: shallow machine learning model-based methods and DNN-based methods. Shallow model-based methods with strong theoretical backgrounds have been proposed since the early 2010s. Scheirer et al. \cite{Scheirer2013} developed a 1-vs-set machine, a variant of linear SVM trained to minimize the open space risk formalized by the authors. In a subsequent work \cite{Scheirer2014}, Scheirer et al. introduced EVT to calibrate the scores of radial basis function (RBF) SVM. They also applied a one-class SVM to supplement the RBF SVM decision scores, providing a theoretical foundation that the CAP model can be derived from a one-class SVM. Recently, the focus of research has been on incorporating the incremental learning of new instances. For example, a nearest non-outlier algorithm that uses thresholded distances from the nearest class mean as its decision function was proposed to incorporate the incremental learning function \cite{Bendale2015}. A nonparametric instance-based model, called the nearest-neighbors distance ratio open set classifier, was developed to reflect the information of new instances \cite{MendesJunior2017}. Unfortunately, these two models lack theoretical grounding. To provide a well-grounded theoretical foundation for both incremental learning and offline decisions, Rudd et al.  \cite{Rudd2018} proposed an extreme value machine, derived from EVT, to perform nonlinear kernel-free incremental learning.

Although DNNs have overwhelmed shallow machine learning models that use manual image features as their input in the field of image recognition, only a few deep models have been proposed for open set recognition. The first deep model introduced for open set recognition was OpenMax, which models the class-specific distribution of activation vectors in the penultimate layer of a CNN and transforms the distributional information into decision scores using an EVT-based calibration strategy \cite{Bendale2016}. Ge et al. \cite{Ge2017} extended this model by incorporating a conditional generative adversarial network (GAN) to generate synthetic samples and to acquire information about unknown classes. Neal et al. \cite{Neal2018} used an encoder-decoder GAN to generate counterfactual images and trained these images as the “unknown” class. However, whether the resulting synthetic samples can represent possible unknown examples remains unknown. In the field of natural language processing, Shu et al. \cite{Shu2017} trained a CNN with a sigmoid output layer and performed score analysis on the final layer activations to obtain a class-specific reject/accept threshold. However, they did not provide a theoretical background for using sigmoid activation scores for post recognition analysis, even though the proposed deep open classification (DOC) network outperformed OpenMax. Recently, Oza and Patel \cite{Oza2019} showed that there is a difference in reconstruction error calculated by class-conditioned autoencoders for match and nonmatch examples. Based on observations, they were generated a class membership probability by introducing generalized Pareto distribution-based calibration. Most of these DNN-based open set recognition methods lack a theoretical foundation for adopting post recognition analysis.

\section{Deep Open Probability Model} \label{model}

\subsection{The Structure and Background of Proposed Model}

We propose a model that combines a convolutional neural network (CNN) feature extractor and the following one-vs-rest sigmoid networks (see  Fig. \ref{structure} in the supplementary section \ref{spplementary}). For each target class, a one-vs-rest sigmoid network is trained to map latent representations of target class examples to one and map those of the other classes to zero. Additionally, the decision scores of one-vs-rest networks are transformed to a class membership probability by the EVT-based calibration method.

The SoftMax function $s(z_y)=\frac{e^{z_y}}{\Sigma_{y\in \Upsilon}e^{z_y}}$, which is the de facto standard activation used for multiclass classification, is the normalized exponential function of the logit $z_y$ for target class $y$. A CNN with SoftMax output layer is trained to increase the relative quantity of $z_y$ to logits of the other classes, which causes the activations of nontarget classes to converge to zero by increasing the normalization term (the denominator), even though the logits of nontarget classes do not decrease substantially \cite{Boult2019,Oland2017}. The activations that converge to zero backpropagate very small gradients through the network and the network rarely learns hidden representations that can be useful for differentiating nonmatch samples. However, the sigmoid function $\rho (z_y)=\frac{1}{1+e^{-z_y}}$ is not affected by the other output nodes. Match and nonmatch examples are equally considered in each sigmoid output node during training. Thus, each sigmoid node is trained to differentiate a dissimilar example from match examples. Naturally, a CNN with a sigmoid output node is more likely to reject unknown from known classes based on dissimilarity than is a CNN with a SoftMax output layer.

In a typical CNN architecture, where a single output layer follows the final hidden layer, only a very simple probability model can be constructed on the feature space. Specifically, if we assume probability estimating function is usually a monotonically increasing function of a sigmoid activation, the membership probability of class $y$ is a simple monotonically increasing function in the positive direction of vector $W_y^{fo}$ in the feature space, where $W_y^{fo}$ is the weight vector that maps the final hidden nodes to the output. This is because sigmoid is a monotonically increasing function of logit. If we set a probability threshold in this probability model, a linearly separated positively labeled area is given to a class. This result is not a problem for general classification tasks because the feature space can be trained to differentiate different known class samples linearly. However, the simple probability model is likely to generate redundant positively labeled space, in which positive samples do not exist. However, the simple probability model is likely to generate redundant positively labeled space, in which positive samples do not exist. However, one-vs-rest sigmoid networks can provide a more sophisticated probability model, because the one-vs-rest network can provide a flexible nonlinear features-to-score mapping for each target class. This flexible mapping can reduce the possibility of containing redundant space as a positive area. An illustrative example comparing simple and sophisticated probability models can be found in our supplementary. Additionally, ReLU activation can be used for the hidden layers to construct a probability model on the feature space, holding the properties of the CAP model in a subspacewise manner. Finally, the derived probability model minimizes and manages open space risk by controlling the size of the open space in the feature space.

\subsection{Open Space Risk}

The well-known definition of open space risk was provided by Scheirer et al. \cite{Scheirer2013}. Let $f$ be a measurable recognition function, where $f_y(h)=1$ for recognition that a sample $h\in \mathcal{H}$ is included in a class $y$ and $f_y(h)=0$ otherwise, where $\mathcal{H}$ is the feature space. They defined open space risk $R_{\mathcal{O}}(f)$ for class $y$ as $R_{\mathcal{O}}(f)=\frac{\int_\mathcal{O}f_y(h)dh}{\int_{S_o}f_y(h)dh}$, where $\mathcal{O}$ is the open space and $S_o$ is a ball of radius $r_o$ that includes all the known positive training examples. In this definition, open space risk is considered to be the relative measure of positively labeled open space compared to the overall measure of positively labeled space. The definition of open space risk requires a definition of open space $\mathcal{O}$. Open space $\mathcal{O}=S_o-\cup_{i=1}^{N}B_r(h_i)$ is defined as the space sufficiently far from any known positive training sample $h_i\in \mathcal{H}, i=1,…,N$, where $B_r(h_i)$ is a closed ball of radius $r$ centered on training sample $h_i$. In this paper, we call $B_r(h_i)$ the plausible positive space around $h_i$.

The rationale behind the closed-ball-shaped plausible positive space is that a probability model in which class membership probability abates as data points move from known data in all directions can be obtained. However, this assumption holds for only a few shallow machine learning models. To incorporate the concept of open space risk into DNNs, we redefine open space and open space risk based on the relationship between the hidden representation and the output of the network. Before presenting the definition, it is assumed that a well-trained neural network that uses ReLU hidden activation and sigmoid output activation follows the DNN feature extractor. In other words, for a class $y$, threshold $\delta$ can be set lower than all positive training samples in the sigmoid output space, while most negative training samples can be left below the threshold. Let $\rho$ be the sigmoid function and $g_y$ be the mapping function of the neural network that produces the logit of class $y$ by inputting $h \in \mathcal{H}$. We define comprehensive positive space $S_p$ as shown in \eqref{eq1}, and open space risk  $R_{\mathcal{O}}(f)$ is modified based on $S_p$ as shown in \eqref{eq2}.
\begin{gather}
S_p=\{h\mid \rho (g_y(h))>\delta , h \in \mathcal{H}\}. \label{eq1}\\
R_{\mathcal{O}}(f)=\frac{\int_\mathcal{O}f_y(h)dh}{\int_{S_p}f_y(h)dh} \label{eq2}
\end{gather}
In this paper, we consider open space as the set of points of which the activations are lower than those of any known training samples in positively labeled area, allowing a soft margin $r$. Thus, the formal definition of open space is provided in \eqref{eq3}, where $\Delta_r(h_i)=\{h\mid \rho(g_y(h))\geq \rho(g_y(h_i))-r, h\in \mathcal{H}\}$.
\begin{gather} 
\mathcal{O}=S_p-\cup_{i=1}^{N}\Delta_r(h_i). \label{eq3}
\end{gather}

\subsection{Compact Abating Probability Modeling}

To construct a probability model that can explain increases and decreases in the probability score in the feature space, the probability model should be based on an abating function. An abating function $A(h)$ is a function that satisfies the property $\forall h,\exists h^* \mid 0<A(h)\leq \Psi(d(h,h^*))$, where $\Psi(d)$ is the upper bound of the abating function, which is a nonnegative finite square-integrable continuous decreasing function of distance measure $d$ \cite{Scheirer2014}. That is, the abating function and bound monotonically decrease as the distance between the positive training sample $h^*$ and any point $h$ increases. Because an abating bound is nonnegative finite square-integrable, an abating function can be transformed into a probability function via calibration. We extend this abating function to the directional abating function $A_W(h)$, which abates in the direction of $W$. $A_W(h)$ satisfies the following inequality:
\begin{gather} 
0<A_W(h)\leq \Psi(W\cdot (h-h^*))\: \exists h^*, \forall h \in U_W(h^*), \label{eq4}
\end{gather}
where $U_W(h^*)=\{ h \mid W\cdot (h-h^*)>0\}$ and $\Psi(W\cdot (h-h^*))$ is the abating bound in the direction of $W$. From \eqref{eq4} and a property of the sigmoid function, we can derive the following.

\textbf{Theorem 1} (The sigmoid activation is a directional abating function). Let $A_{-W}(h)=\rho(W\cdot h)$ for $h\in \mathcal{H}$; then, $A_{-W}(h)$ is a directional abating function in the direction of $-W$.

\emph{Proof.} Let $h_i\in \mathcal{H}, i=1,…,N$, be the feature vectors for the training dataset of class $y$, let $M$ be a sufficiently large number that satisfies $\forall i \mid M\Vert W\Vert^2\gg W\cdot h_i$, and let $h^*$ be a positive training feature vector. We define $\Psi(d(h,h^*))=\rho(-d(h,h^*)+ M\Vert W\Vert^2)$, where $d(h,h^*)=-W\cdot (h-h^* )$ for$ h\in U_{-W}(h^* )$. Then, by definition,$\Psi(d(h,h^*))=\rho(W\cdot h+(M\Vert W\Vert^2-W\cdot h^*))$ is always larger than $\rho(W\cdot h)$. Moreover, because sigmoid $\rho$ is a monotonically increasing function with $\rho(z)>0$ for all $z\in \mathbb{R}$, $\Psi(d(h,h^*))$ is decreasing as the distance $d(h,h^*)$ increases. Additionally, $\Psi(d(h,h^*))$ is a nonnegative finite square-integrable continuous decreasing function according to \eqref{eq5}. Thus, \eqref{eq4} is always satisfied for $A_{-W}(h)$ and $\Psi(d(h,h^*))$.
\begin{gather} 
\int\limits_0^\infty \Psi(d(h,h^*))^2 dd(h,h^*)<\int\limits_0^\infty \Psi(d(h,h^*)) dd(h,h^*)=\int\limits_{-\infty}^{M\Vert W\Vert^2} \rho(z) dz=ln(1+e^{M\Vert W\Vert^2})  \label{eq5}
\end{gather}

Neural networks that use ReLU hidden activation layers decompose the input space into multiple polyhedra, and on each polyhedron, the output logit and features have a linear relationship \cite{He2018}. Without loss of generality, Theorem 1 holds for any polyhedron in the feature space. Let $\mathcal{H}=\cup_{j=1}^{K}\Omega_j$ , where $\Omega_j$ is a polyhedral subspace decomposed by the following neural network that applies ReLU hidden activation and $K$ is the number of subspaces. Then, the subspacewise directional abating function $A_{-W_{\Omega_j}}(h\mid h\in \Omega_j )=\rho(W_{\Omega_j}\cdot h)$ is defined for each subspace $\Omega_j$, where $W_{\Omega_j}$ is the weight vector that produces the logit for $h \in \Omega_j$.

Sigmoid activation has been used to estimate the probability of class membership in machine learning models \cite{Zadrozny2002,Platt2000}. However, because a DNN with a sigmoid output layer, unlike a DNN with a SoftMax output layer, is highly likely to overfit training examples, calibration is needed to estimate the probabilities of both known and unknown examples. Thus, we define the probability model $P$ that calibrates the sigmoid output of the DNN for class $y$ as follows:
\begin{gather} 
P(y\mid h\in \Omega_j)=C(\rho(g_y(h))), \label{eq6}
\end{gather}
where $C$ is a monotonically increasing calibration function. We provide the EVT-based calibration function in Section \ref{estimation}. The probability model defined in \eqref{eq6} is converted to the following conditional probability model:
\begin{gather} 
P(y\mid h\in \Omega_j)=C(A_{-W_{\Omega_j}}(h))\: \forall j \in \{1, 2, ...,K\}. \label{eq7}
\end{gather}
This conditional probability model is based on the subspacewise directional abating function; thus, we call the model the subspacewise directional abating probability model.

The subspacewise directional abating probability model can provide a low probability for points in the open space of the subspace. However, there is still the risk that unknown examples are recognized as belonging to a known class because the model can have nonzero probability over open space $\mathcal{O}$. To address this problem, we define a subspacewise directional CAP model with a sigmoid activation threshold $\tau$, and the model $P_\tau$ satisfies the following property:
\begin{gather} 
\rho(W_{\Omega_j}\cdot h)<\tau \Rightarrow P_{\tau}(y\mid h\in \Omega_j)=0\: \forall j \in \{1, 2, ...,K\}. \label{eq8}
\end{gather}
From the definition of the subspacewise directional CAP model, we can derive the following core theorem. Thus, the proposed one-vs-rest networks can be used as the base model to yield the class membership probability by managing open space risk.

\newcommand{\argmin}{\arg\!\min}
\textbf{Theorem 2} (A subspacewise directional CAP model can minimize and manage open space risk). Let $P_{\tau}(y\mid h\in \Omega_j),\: j=1,…,K$ be subspacewise directional CAP models and $f_y(h)$ be a measurable function of $h \in \mathcal{H}$ that determines the class membership based on $P_{\tau}(y\mid h\in \Omega_j)$. Let $ R_{\mathcal{O}}(f)$ be the open space risk and $\mathcal{O}$ be open space, defined as in \eqref{eq2} and \eqref{eq3}, respectively. If $\tau$ in \eqref{eq8} satisfies $\max_j \rho(W_{\Omega_j}\cdot h_{\Omega_{j^*}})-r<\tau$, where $h_{\Omega_{j^*}}=\argmin_{h_i\in \Omega_j}\rho(W_{\Omega_j}\cdot h_i)$ and $h_i \in \mathcal{H}, i=1,…,N$, are the feature vectors for the training dataset of a class $y$, then $R_{\mathcal{O}}(f)=0$. That is, the open space risk of subspacewise directional CAP models can be zero. From this and the property of the CAP defined in \eqref{eq8}, we derive that open space risk can be managed by adjusting $\tau$.

\emph{Proof.}
Let $\mathcal{O}'=S_o-\cup_{i=1}^{N}\Delta_{\Omega(h_i),r}(h_i)$, which satisfies $\mathcal{O}\subseteq \mathcal{O}'$, where $\Omega(h_i)$ is the polyhedral subspace containing $h_i$ and $\Delta_{\Omega(h_i),r}(h_i)=\{h\mid \rho(W_{\Omega(h_i)}\cdot h)\geq \rho(W_{\Omega(h_i)}\cdot h_i)-r,h \in \Omega(h_i)\}$. $\mathcal{O}'$ can be reformulated as $\mathcal{O}'=S_p-\cup_{j=1}^{K}\Delta_{\Omega_j,r}(h_{\Omega_j^*})$. According to this definition and the definition of $S_p$ in \eqref{eq1}, $\mathcal{O}'$ is reformulated as:
\begin{gather} 
\mathcal{O}'=\cup_{j=1}^{K}\{h\mid \delta \leq \rho(W_{\Omega_j}\cdot h)<\rho(W_{\Omega_j}\cdot h_{\Omega_j^*})-r,\: h \in \Omega_j\}. \label{eq9}
\end{gather}
Since $\forall j\mid \rho(W_{\Omega_j}\cdot h_{\Omega_j^*})-r<\tau$ and $P_\tau (y\mid h\in \Omega_j)=0$ for $\rho(W_{\Omega_j}\cdot h)<\tau$, we have $\int_{\mathcal{O}'}f_y(h)dh=0$. From $\mathcal{O}\subseteq \mathcal{O}'$,  $\int_{\mathcal{O}}f_y(h)dh=0$ is satisfied, which yields $R_\mathcal{O}(f)=0$.

\subsection{Probability Estimation and Open Set Recognition} \label{estimation}

The remaining issue that to address to complete the deep probability model is estimating class membership probability by applying a calibration method, as shown in \eqref{eq7}. Because there are no unknown examples during training, nonmatch examples are used as substitutes for unknown examples. The main hurdle of open set recognition is in differentiating similar match and nonmatch examples. In other words, when constructing an open set recognition system, the extreme examples are more informative than the examples with scores in the center area of a particular class score distribution \cite{Scheirer2012}. Additionally, the extreme values of score distribution produced by a well-trained recognition model can be modeled by an EVT distribution, which provides probability estimates for match and nonmatch samples \cite{Scheirer2013,Oza2019,Scheirer2011,Gibert2015}. Therefore, we applied the EVT-based calibration technique to the sigmoid outputs.
According to the statistical EVT, the extreme values of a score distribution follow a Weibull distribution if the scores are bounded below and follow a reverse Weibull distribution if bounded above \cite{Kotz2001,NIST2008}. We construct two probability functions that estimate the probability of “being a positive” and the probability of “not being a negative”. Before modeling, we first separate training examples into match examples and nonmatch examples for class y. Let us assume that $s_i^y=F_y(x_i)$ is the score of example $x_i$ for class $y$, where $F_y$ is the mapping function of the proposed model that produces sigmoid output for class $y$. Then, we can collect the scores of matches $S_{y+}$ and the scores of nonmatches $S_{y-}$ separately. Let $E_{y+}$ be the lower extremes of the matches $S_{y+}$ and $E_{y-}$ be the upper extremes of the nonmatches $S_{y-}$. The relative quantity of extremes of matches and nonmatches is represented by the ratio parameter $\alpha$. Because $E_{y+}$ follows a Weibull distribution, we use the Weibull CDF to estimate the probability of being a positive, as shown in \eqref{eq10}. Furthermore, the reverse Weibull CDF is used to estimate the probability of not being a negative, as shown in \eqref{eq11}. The Weibull and reverse Weibull distributions have three parameters: location $\nu$, scale $\lambda$, and shape $\kappa$. We estimate $\nu_{y+}$, $\lambda_{y+}$, and $\kappa_{y+}$ that best fit $E_{y+}$ and $\nu_{y-}$, $\lambda_{y-}$, and $\kappa_{y-}$ that best fit negative values of $E_{y-}$ by applying the maximum likelihood method provided by the SciPy library of Python \cite{Jones2001}.
\begin{gather} 
P_+(y \mid F_y(x_i))=1-e^{-(\frac{F_y(x_i)-\nu_{y+}}{\lambda_{y+}})^{\kappa_{y+}}}. \label{eq10}\\
P_-(y \mid F_y(x_i))=e^{-(\frac{-F_y(x_i)-\nu_{y-}}{\lambda_{y-}})^{\kappa_{y-}}}. \label{eq11}
\end{gather}
Finally, we propose a recognition rule to assign a sample to a known class or to reject it. We use the product $P_+(y \mid F_y(x_i))\times P_-(y \mid F_y(x_i))$ as $P(y \mid F_y(x_i))$, the probability of belonging to class y. That is, if an example has a high probability of being from the matches and not from nonmatches, then the example is highly likely to belong to the positive class. However, we reject an example if it cannot be considered to belong to any of the known classes. Thus, the multiclass open set recognition rule is defined as follows:
\newcommand{\argmax}{\arg\!\max}
\begin{gather} 
y^*= \begin{cases}
    \argmax_{y\in \Upsilon}P(y \mid F_y(x_i)) & \text{if $P(y \mid F_y(x_i))\geq\theta$}\\
    \text{unknown} & \text{otherwise},
  \end{cases}. \label{eq12}
\end{gather}
where $\theta$ is the lower limit to be assigned to one of the known classes $y\in\Upsilon$, which is the calibrated value of $\tau$ in \eqref{eq8}. We select the optimal $\theta$ and $\alpha$, the ratio of extremes, that maximize the open set recognition performance via cross-class validation \cite{Bendale2015}, which measures the performance for the validation set while leaving out a randomly selected subset of known classes as “unknown”.

\section{Experiments} \label{experiments}

We conducted several experiments to demonstrate the effectiveness of the proposed method in open set recognition scenarios. The proposed method was evaluated from two perspectives. The first was whether the proposed model can provide more informative hidden features for unknown classes compared to a CNN with a SoftMax output layer. The second question was whether the proposed method outperforms state-of-the-art CNN-based open set recognition methods. For the first experiment, we assumed that the larger the difference between the distributions of a known class and an unknown class in the final hidden representation space is, the more information we can infer about the unknown class. We designed a new evaluation method comprising three steps. First, we discretized the feature space based on the distributional information of samples belonging to a known class $y$. For each discretized area, the ratio of samples belonging to the area is calculated for the known class and an unknown class $\chi$. We used the K-means clustering algorithm \cite{MacQueen1967} for discretization, and a sample was considered to belong to the cluster area of which the centroid was closest to the sample. Next, the difference in the probability distributions was calculated based on Kullback-Leibler divergence (KL) \cite{Kullback1951} given by:
\begin{gather} 
KL(q_{\chi,y}\parallel q_{y,y})=\sum_i q_{\chi,y}(i)ln(\frac{q_{\chi,y}(i)}{q_{y,y}(i)}), \label{eq13}
\end{gather}
where $q_{a,b}$ is the probability distribution of class $a$, which is calculated based on cluster information of class $b$, and $q_{a,b}(i)$ is the probability for the discretized area $i$. The first and second steps were applied to all pairs of known classes and unknown classes. Finally, we conducted a paired $t$ test and found that the KL distribution of the proposed method is significantly larger than that of a CNN with SoftMax output layer.

We also compared the proposed method with CNN-based benchmarks in terms of multiclass open set classification performance. The following four methods were selected for comparison. OpenMax uses the distance from the mean activation vector of a known class as the decision score, adopting Weibull-based calibration \cite{Bendale2016}. G-OpenMax incorporates conditional GAN to generate synthetic samples into OpenMax \cite{Ge2017}. Open set recognition using counterfactual images (OSRCI) introduces counterfactual images generated by an encoder-decoder GAN and uses these images as training data for the unknown class when learning a CNN \cite{Neal2018}. DOC uses the sigmoid output of a CNN as the decision score and sets the class-specific reject/accept threshold based on a Gaussian distribution \cite{Shu2017}. We report the results in terms of the F-measure, which is commonly used to measure the performance of open set recognition methods \cite{Zhang2017,Scheirer2013}. To diversify the experimental conditions, we dynamically set the openness, which quantifies how open the problem setting is:
\begin{gather} 
openness=1-\sqrt{\frac{2\times NC_T}{NC_E+NC_R}}, \label{eq14}
\end{gather}
where $NC_T$ is the number of classes used in training, $NC_E$ is the number of classes used in evaluation (testing), and $NC_R$ is the number of classes to be recognized.

We performed the evaluation using three datasets: extended MNIST (EMNIST) \cite{Cohen2017}, labeled faces in the wild (LFW) \cite{Huang2007}, and Cifar-100 \cite{Krizhevsky2009}. EMNIST is an extension of the well-known MNIST handwritten digits to commonly used characters, such as the English alphabet. LFW, one of the most popular face recognition benchmark datasets, contains face images of 5,749 individuals. Finally, Cifar-100 is a famous three-channel color image dataset with 100 natural image categories. To ensure consistent comparisons, we used the same base CNN structure and training settings, except the output layer for each dataset, for all comparison methods. Specifically, the model structure and training settings of \cite{Shu2018}, those of \cite{Oza2019}, and the structure of All-CNN-C and the training settings of \cite{Springenberg2014} were used for EMNIST, LFW, and Cifar-100, respectively. In addition, we applied the same base CNN structure to both the encoder and the decoder of the encoder-decoder GAN for OSRCI and to the discriminator of the conditional GAN for G-OpenMax.

\subsection{Open Set Recognition on EMNIST}

\begin{wrapfigure}{r}{0.5\textwidth}
	\vspace{-10pt}
\begin{minipage}[b]{\textwidth}
\begin{subfigure}[b]{0.25\textwidth}
	\includegraphics[width=\textwidth]{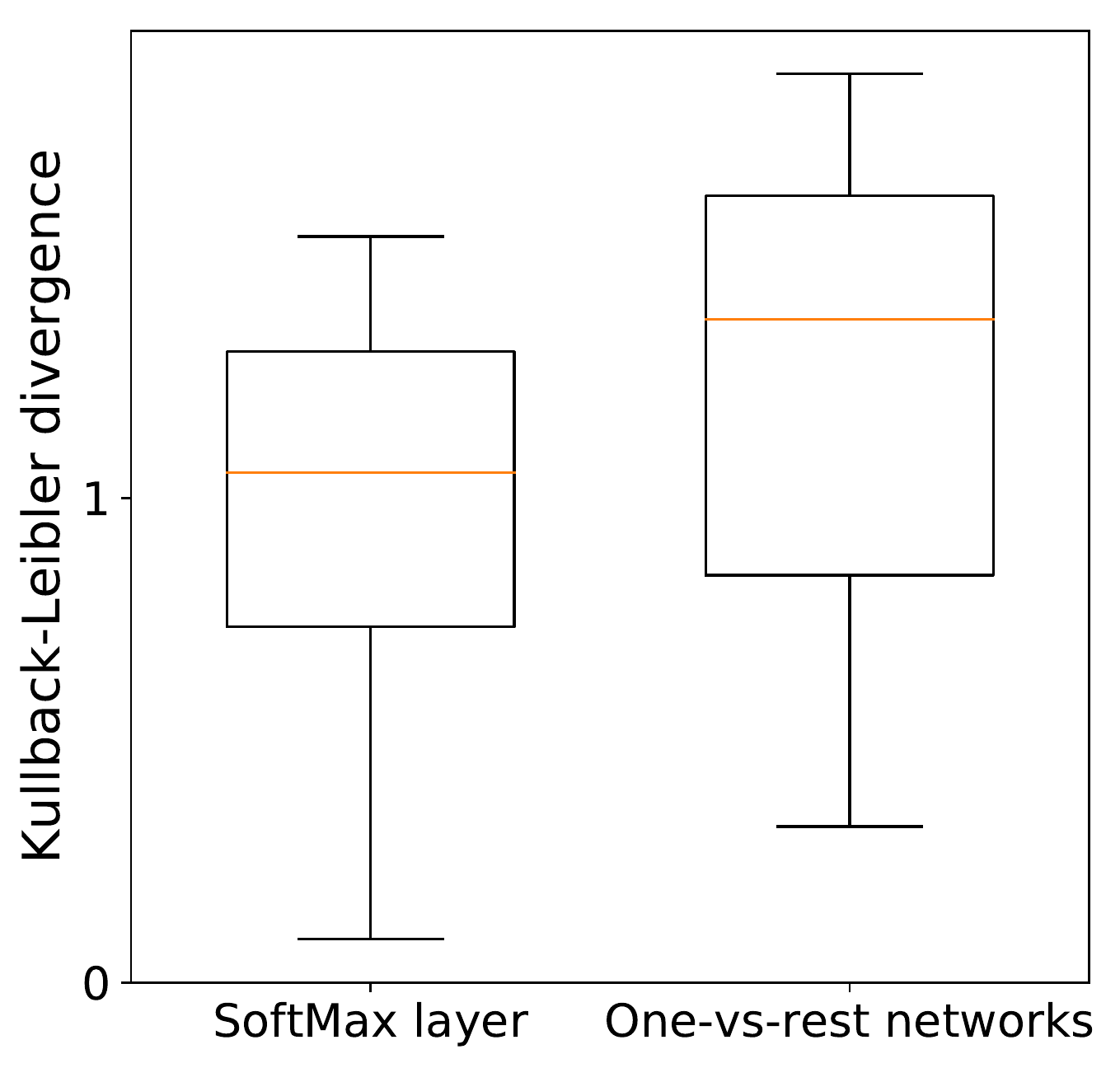}
\subcaption{10 areas}
\end{subfigure}
\begin{subfigure}[b]{0.25\textwidth}
\includegraphics[width=\textwidth]{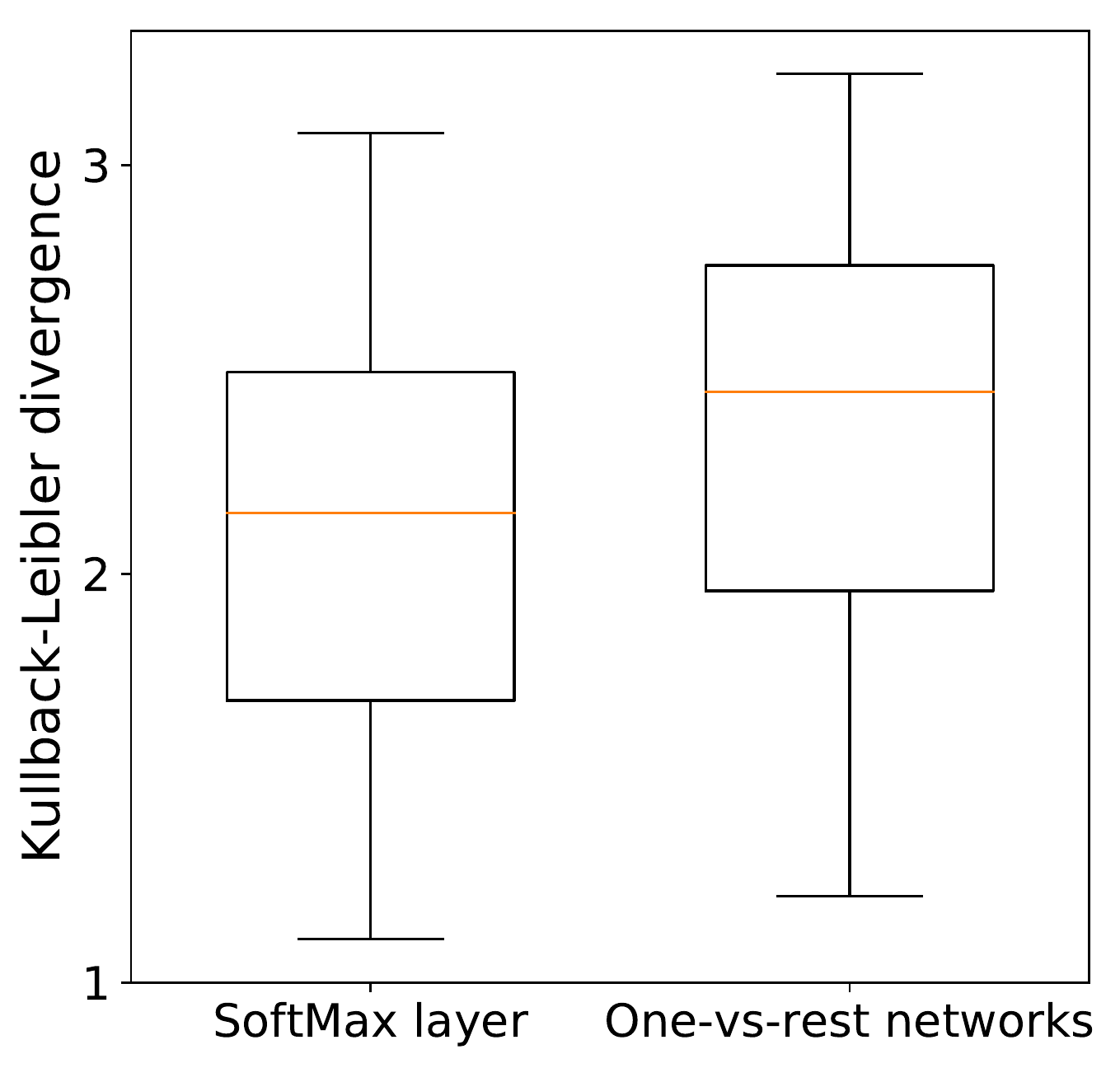}
\subcaption{30 areas}
\end{subfigure}
\end{minipage}
\begin{minipage}[b]{\textwidth}
\begin{subfigure}[b]{0.25\textwidth}
	\includegraphics[width=\textwidth]{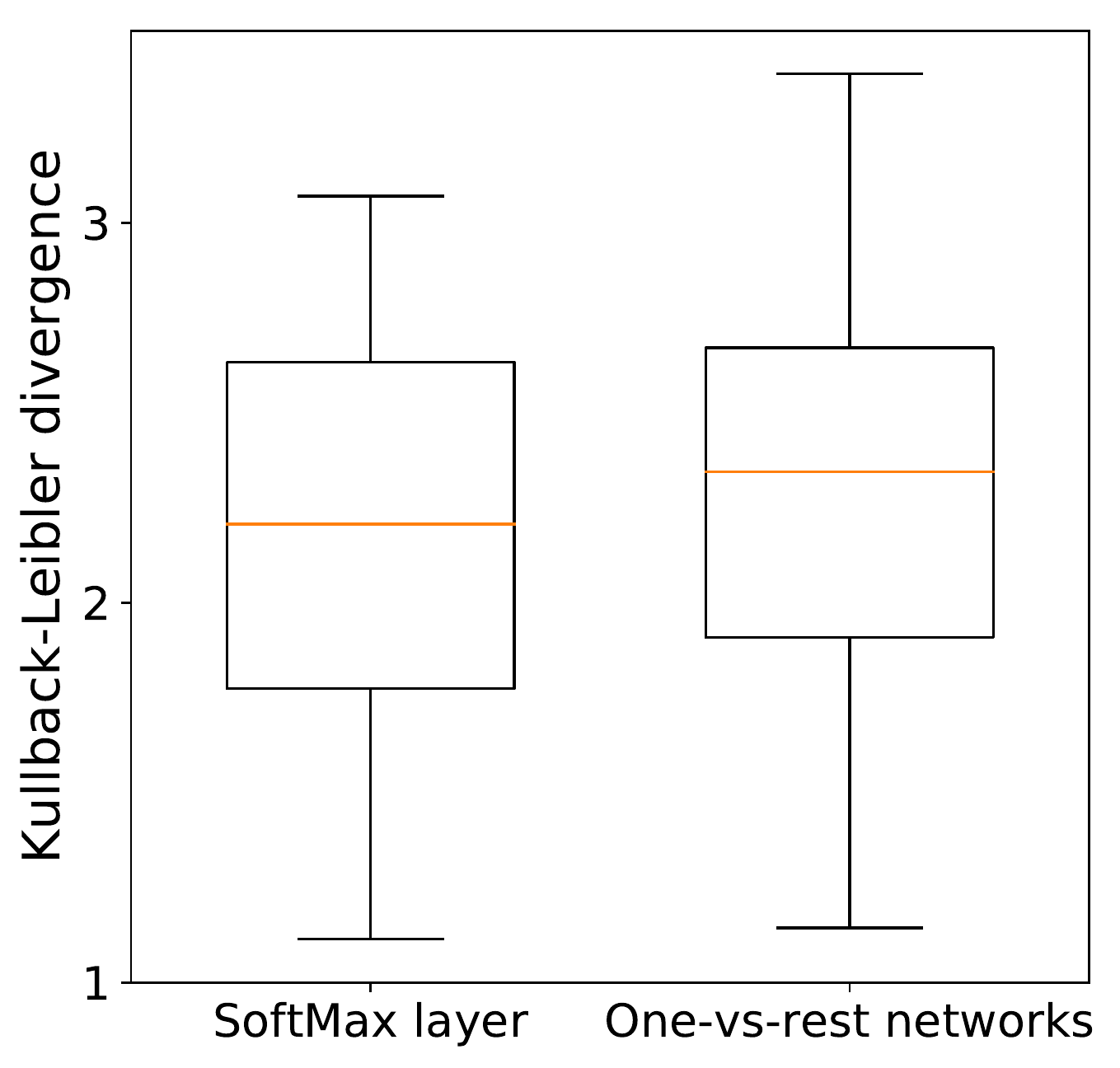}
\subcaption{50 areas}
\end{subfigure}
\begin{subfigure}[b]{0.25\textwidth}
\includegraphics[width=\textwidth]{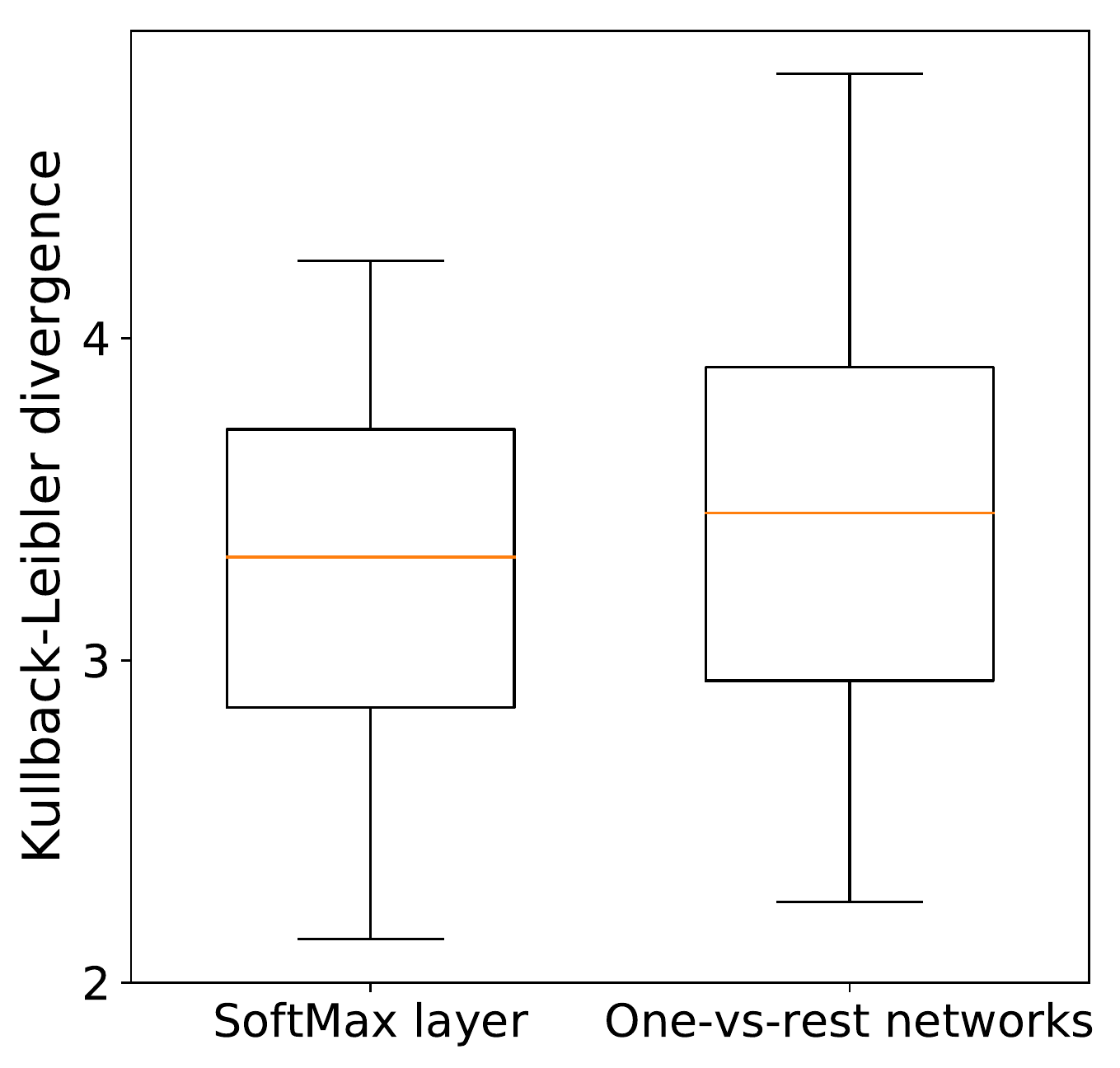}
\subcaption{70 areas}
\end{subfigure}
\end{minipage}
	\caption{Box-and-whisker plot showing the distributions of the KL divergences. The boxes enclose the $25^{th}$ and $75^{th}$ percentiles; the whiskers extend to the $5^{th}$ and $95^{th}$ percentiles; and the orange line is the median.}
	\vspace{-10pt}
  \label{box_and_whisker}
\end{wrapfigure}

To measure and compare the KL divergences, we follow the settings described in the supplementary section \ref{spplementary}. When the feature space is discretized into 10, 30, 50 and 70 areas, the KL divergences for 370 known and unknown class pairs are distributed as shown in Fig. \ref{box_and_whisker}. The results show that the finer the discretized feature space is, the greater the difference between the probability distributions of the known and unknown classes. Additionally, the KL divergence distribution of the one-vs-rest networks is always significantly higher than that of the SoftMax layer. The p-values of the paired $t$ test were less than $10^{-4}$ for all cases

\begin{figure}
\begin{minipage}{0.43\textwidth}
	\includegraphics[trim={0.25\textwidth} 0 {0.5\textwidth} {0.3\textwidth},clip,width=\textwidth]{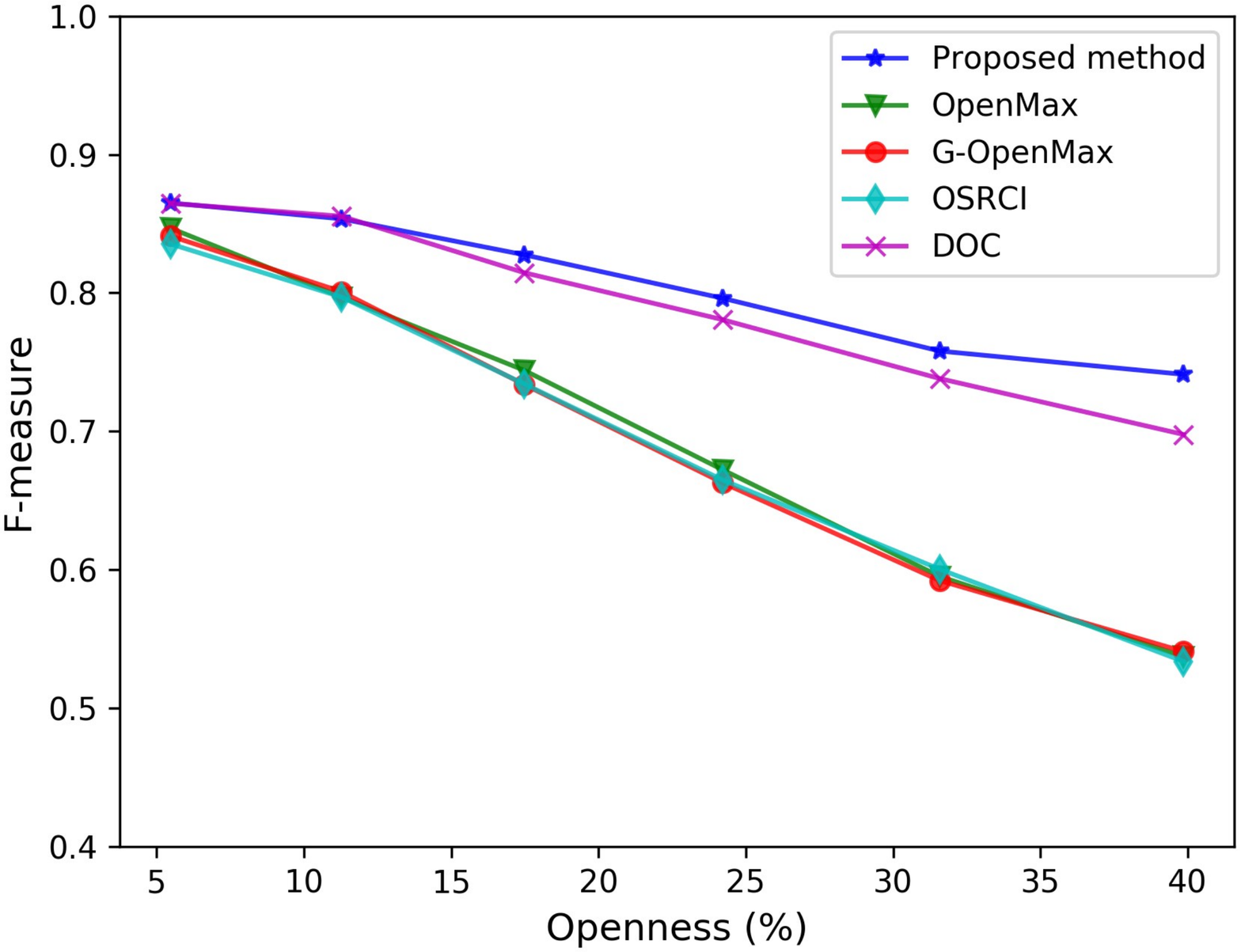}
\caption{F-measure according to openness for open set recognition on EMNIST.}
\label{EMNIST_performance}
\end{minipage}
\hspace{0.05\textwidth}
\begin{minipage}{0.52\textwidth}
\includegraphics[trim={0.5\textwidth} {0.5\textwidth} {0.55\textwidth} {0.55\textwidth},clip,width=\textwidth]{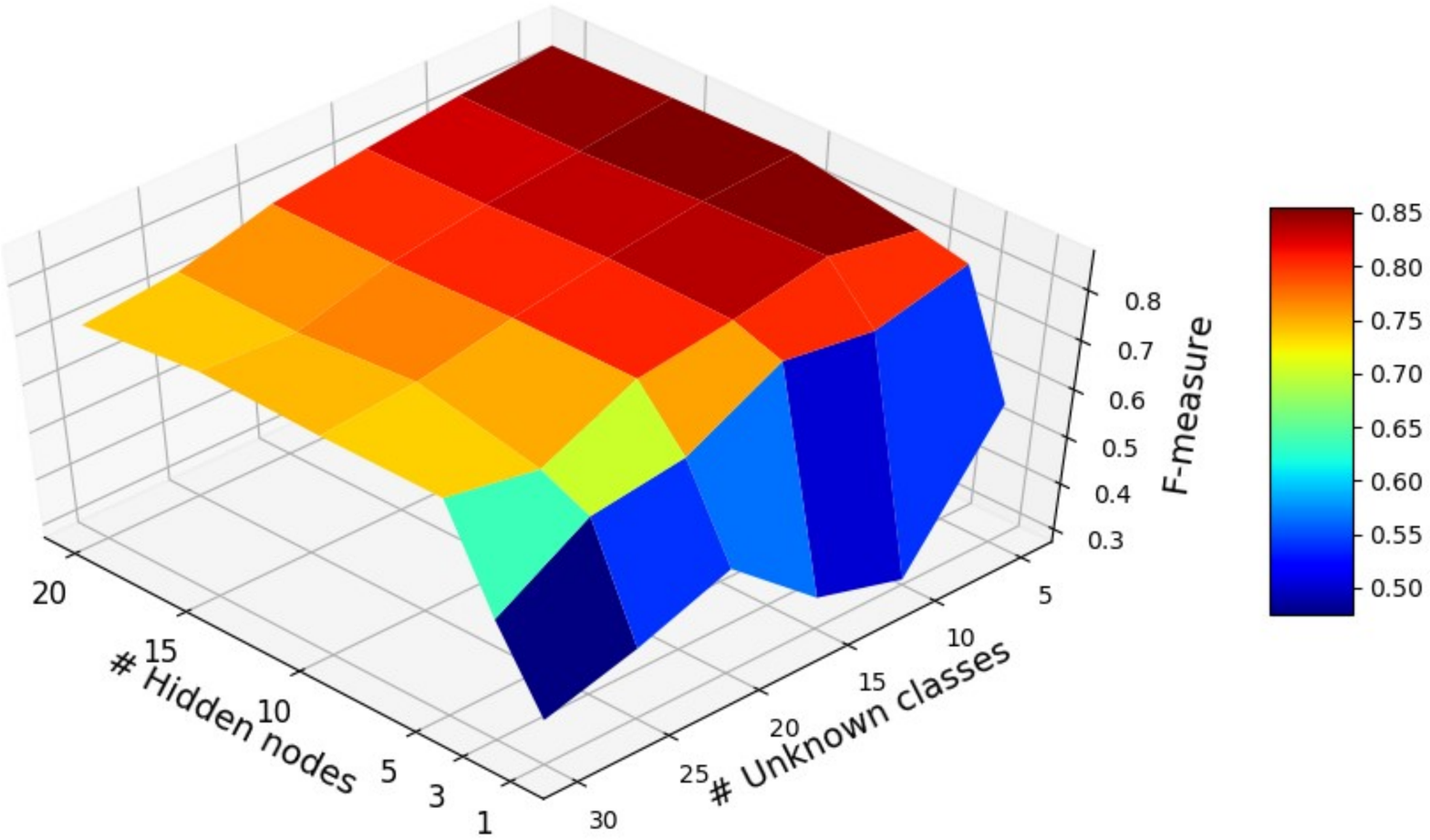}
\caption{Performance evaluation of the proposed method according to the number of unknown classes and the number of hidden nodes.}
\label{EMNIST_parameters}
\end{minipage}
	\vspace{-10pt}
\end{figure}

The proposed method and the four benchmarks were compared under various openness values. To vary the openness, we increased the number of known classes from 5 to 30 in intervals of 5 and randomly selected known classes. The remaining classes were assigned as unknown classes and were not used in training. More details on experimental settings are given in our supplementary. Fig. \ref{EMNIST_performance} shows the evaluation results according to openness and reveals that the proposed method outperformed the other methods as openness increased. Additionally, sigmoid CNN-based methods (the proposed method and DOC) outperformed the SoftMax CNN-based methods in all cases. The performance gap between sigmoid CNN-based methods and SoftMax CNN-based methods increased as the openness increased. This result demonstrates that sigmoid activation can provide a better explanation of unknown examples compared to SoftMax activation. Also, introducing one-vs-rest networks and EVT-based calibration is effective because the F-measure gap between the proposed method and DOC is explained mainly by the output layer structure and the method of estimating class membership probabilities.

We additionally observed the performance differences according to the number of nodes in the hidden layer of the one-vs-rest network and the number of unknown classes, and the results are presented in Fig. \ref{EMNIST_parameters}. For the cases in which the number of hidden nodes was larger than five, there was no substantial difference in the F-measure for the same number of unknown classes. However, as the number of hidden nodes decreased from five, the performance abruptly decreased substantially. Thus, the proposed method should be based on probability estimation through one-vs-rest networks with a sufficiently high level of complexity.

\subsection{Open Set Recognition on LFW}

To evaluate the proposed and comparison methods on the LFW face dataset, we followed the data setting protocol introduced in \cite{Oza2019}. More details on experimental settings are given in our supplementary. We used the training dataset of the known classes and ten largest unknown class datasets to estimate KL divergence distributions. Since the $10^{th}$ unknown class included only 37 images, the feature space was discretized into four to ten areas in intervals of two. As a result, the CNN with one-vs-rest networks always provided larger KL divergences than the SoftMax CNN, yielding p-values less than 0.01 in all cases. The KL divergence distributions according to the number of discretized areas can be seen in supplementary materials. 

\begin{figure}[h]
\begin{minipage}{0.47\textwidth}
	\includegraphics[trim={0.15\textwidth} 0 {0.18\textwidth} 0, clip,width=\textwidth]{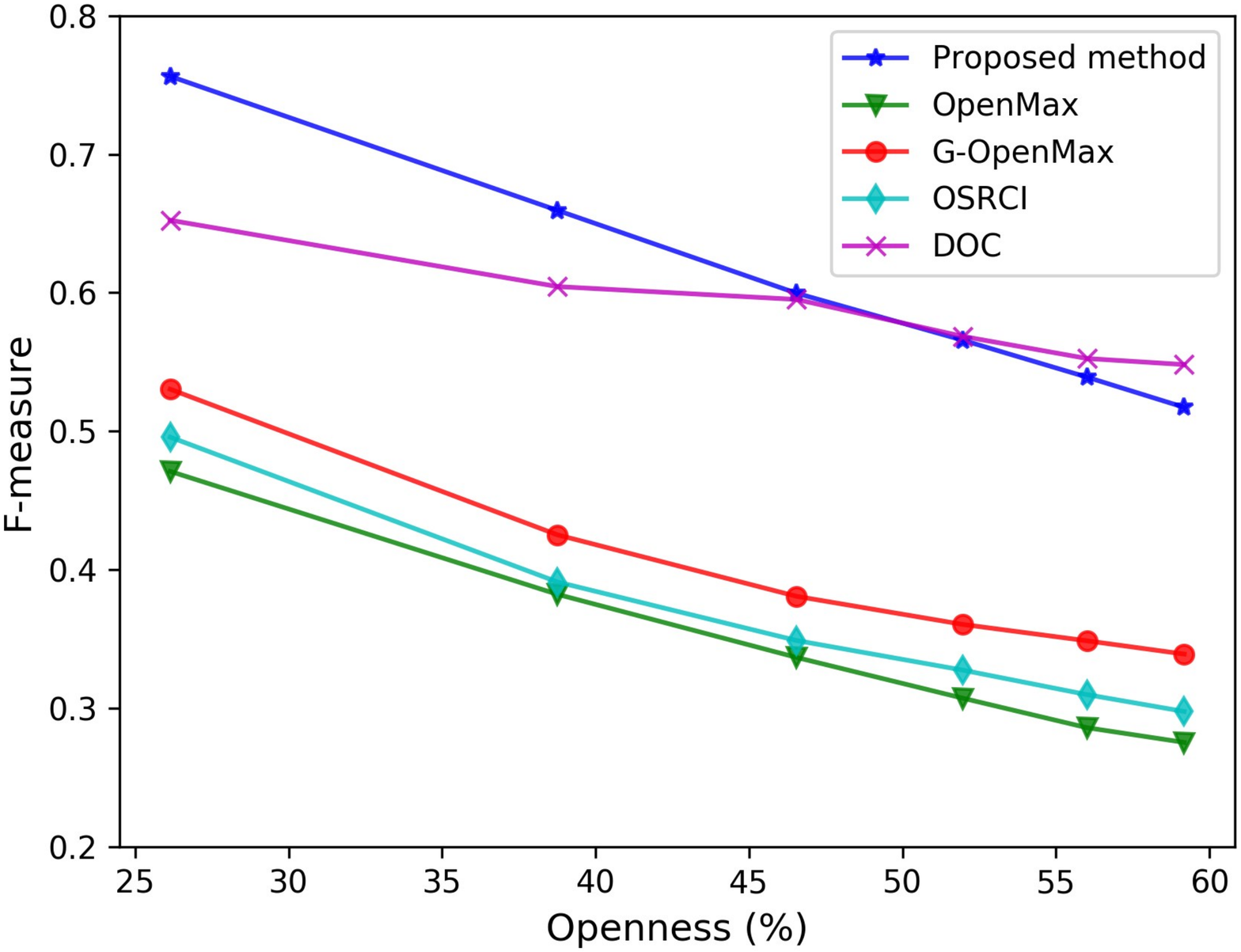}
\caption{F-measure according to openness for open set recognition on LFW.}
\label{LFW_performance}
\end{minipage}
\hspace{0.05\textwidth}
\begin{minipage}{0.47\textwidth}
\includegraphics[trim={0.25\textwidth} {0.05\textwidth} {0.15\textwidth} {0.15\textwidth},clip,width=\textwidth]{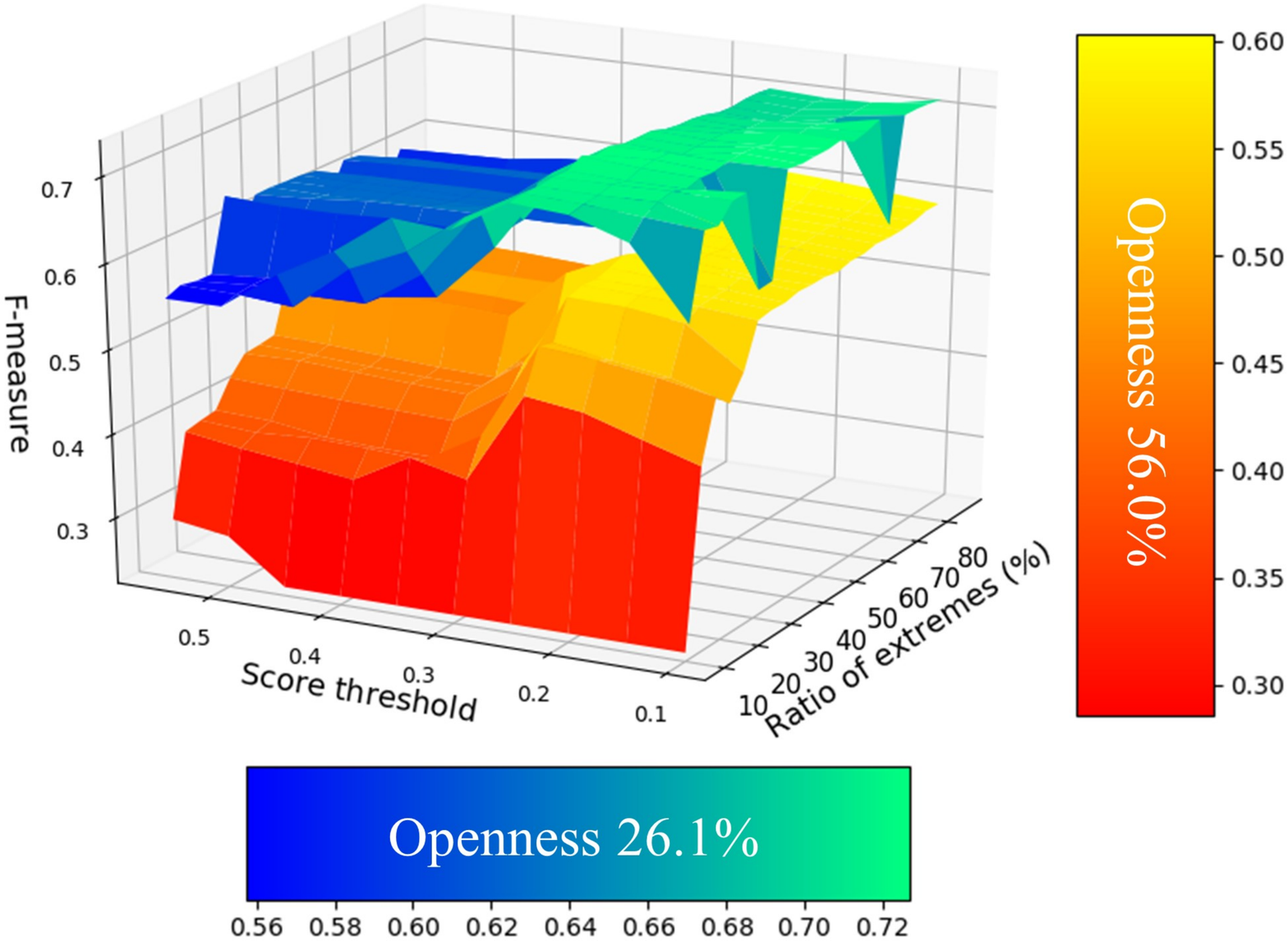}
\caption{Performance evaluation of the proposed method according to the hyperparameters.}
\label{LFW_hyperparameters}
\end{minipage}
	\vspace{-10pt}
\end{figure}

We varied openness from $26\%$ to $59\%$ by taking 10 to 60 unknown classes to evaluate the classification performance. Fig. \ref{LFW_performance} shows the performance evaluation results. The results are the same as those for EMNIST in that the proposed method and DOC outperformed the SoftMax-based methods. Additionally, DOC slightly outperformed the proposed method when the openness exceeded $50\%$. However, the proposed method provided a much higher F-measure at lower openness. Considering that an openness greater than $50\%$ means there are at least three times more unknown than known classes, which is uncommon in the real world, providing better performance at moderate openness using the proposed method is realistic and acceptable.

Fig. \ref{LFW_hyperparameters} shows the performance of the proposed method according to $\theta$ and $\alpha$. For the LFW dataset, $\theta$ had a greater impact on performance than $\alpha$, and performance was more affected by $\theta$ when openness was large. In regard to $\alpha$, it did not have a substantial impact on performance in the case of low openness but had a considerable impact in the case of high openness. Specifically, a small tail size can deteriorate performance in the case of high openness, because seven of the known classes had fewer than one hundred samples and the quantity of data was insufficient to fit the Weibull distribution well when a low ratio of extremes was applied. The poorly fit Weibull distribution could not provide accurate probability estimates.

\subsection{Open Set Recognition on Cifar-100}

We randomly selected 40 classes from a total of 100 classes as known classes, and the remaining classes were considered unknown. Cifar-100 contains 500 training images and 100 testing images for each class \cite{Krizhevsky2009a}. For one-vs-rest networks, a single hidden layer of ten nodes is applied. We used the training images to attain KL divergence distributions. The feature space was discretized into 5 to 30 areas in intervals of 5. For all cases, the paired t test revealed that the one-vs-rest networks provided a significantly larger difference in the distributions of known class samples and unknown class samples in the feature space, with a p-value less than $10^{-25}$. The KL divergence distributions according to the number of discretized areas can be seen in supplementary materials. 

\begin{wrapfigure}{r}{0.45\textwidth}
	\vspace{-10pt}
	\includegraphics[trim={0.15\textwidth} {0.1\textwidth} {0.25\textwidth} {0.10\textwidth},clip,width=0.43\textwidth]{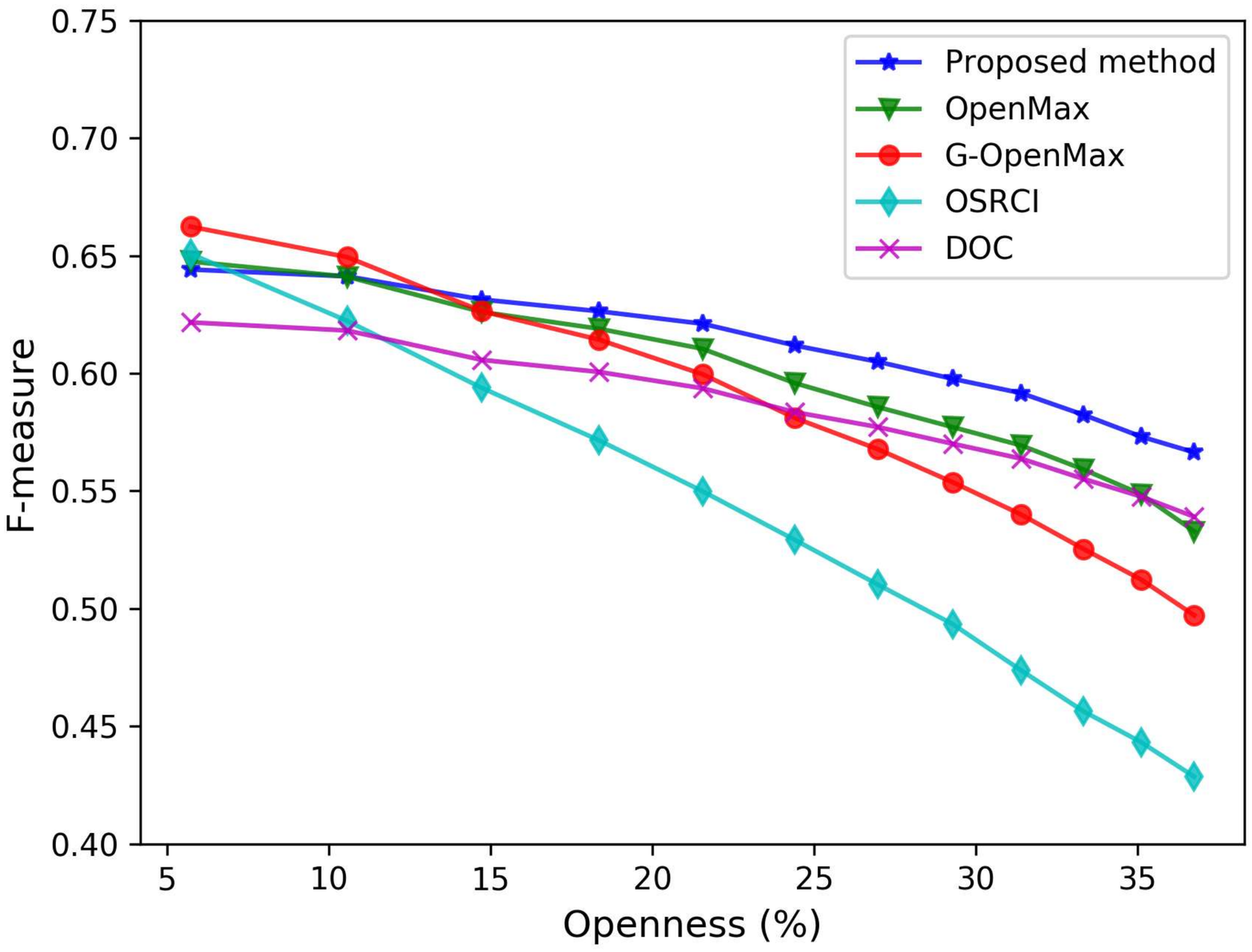}
	\caption{F-measure according to openness for open set recognition on Cifar-100.}
	\vspace{-10pt}
  \label{Cifar_performance}
\end{wrapfigure}

We varied openness from $6\%$ to $37\%$ by taking 5 to 60 unknown classes to compare the performances of the open recognition methods. We applied cross-class validation, leaving ten training classes as unknown classes to find the optimal $\theta$ and $\alpha$. The comparison results are summarized in Fig. \ref{Cifar_performance}. When openness was $5.7\%$ and $10.6\%$, the SoftMax-based methods outperformed the proposed method, but the performance gap was small. However, as openness increased, the proposed method significantly outperformed the other methods. These results can be explained by the observation that the proposed one-vs-rest networks provide more informative latent features for unknown examples.

\begin{figure}[h]
\centerline{
\begin{subfigure}[b]{0.47\textwidth}
	\includegraphics[width=\textwidth]{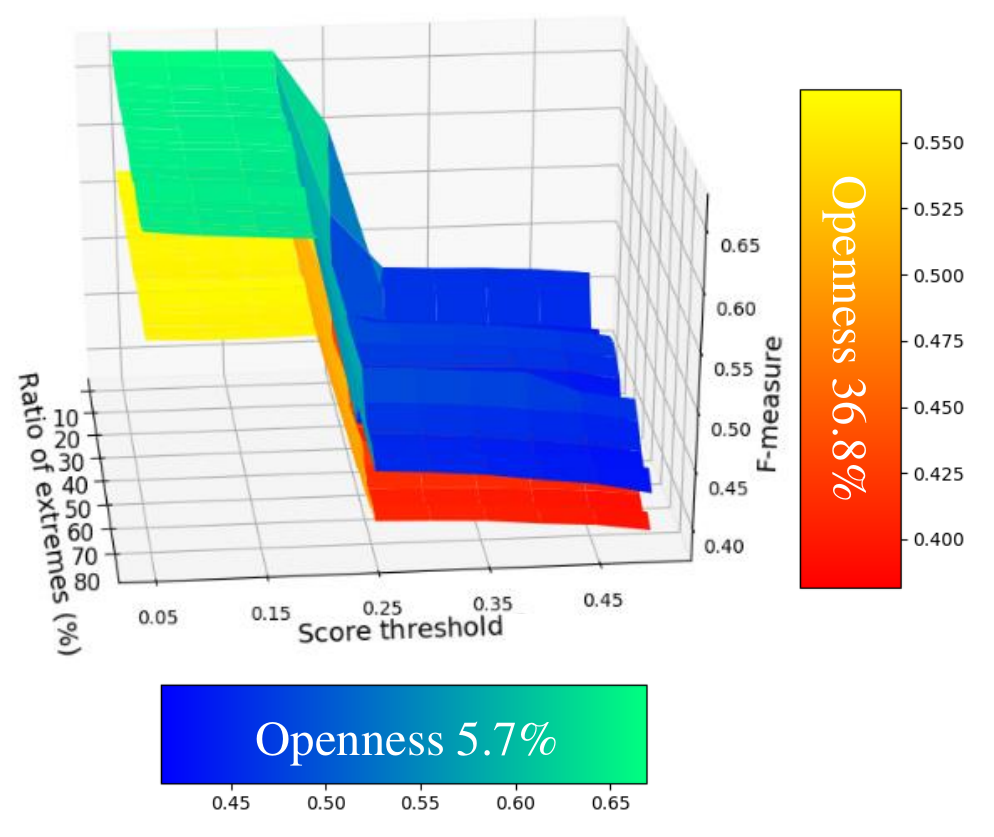}
\subcaption{$\theta \in [0.05, 0.5]$}
\end{subfigure}
\hspace{0.05\textwidth}
\begin{subfigure}[b]{0.47\textwidth}
\includegraphics[width=\textwidth]{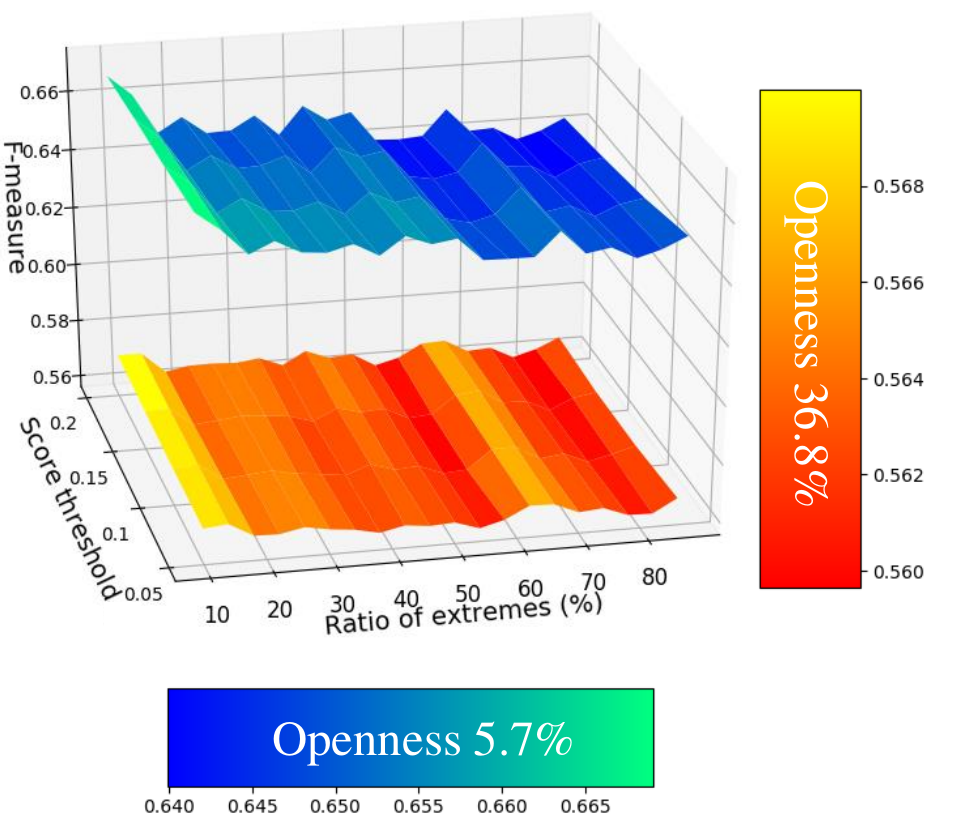}
\subcaption{$\theta \in [0.05, 0.2]$}
\end{subfigure}
}
	\caption{Performance evaluation of the proposed method according to hyperparameters.}
	\vspace{-10pt}
\label{Cifar_hyperparameters}
\end{figure}

We set various $\theta$ and $\alpha$ and observed the performance differences according to the parameter combinations, as shown in Fig. \ref{Cifar_hyperparameters}. Overall, performance determined mostly by the score threshold. However, there is no substantial difference in the F-measure when $\theta$ is less than 0.2. Notably, when the score threshold is lower than 0.2, the minimum ratio of extremes yields the best F-measure. This explains why a small quantity of extremes should be used to calibrate decision scores when the quantity of data is sufficient.

\section{Conclusion} \label{conclusion}

In this paper, we have shown that the proposed one-vs-rest sigmoid networks have powerful ability to learn nonmatch examples. Thus, the proposed model was powerful in differentiating the distributions of known and unknown samples in the feature space. We have also shown that strong theoretical ground exists for estimating class membership probability under the open set scenario when one-vs-rest sigmoid networks follow a DNN feature extractor. Thus, we have provided more than just a high-performance open set recognition method.

For future work, we are interested in extending the proposed model for training in online and incremental manner. When the model is trained with a newly observed unknown sample, the estimated information about the unknown sample would guide and support the training. Additionally, as a second possible future work, we will consider incorporating the concept of open space risk minimization into DNN training to provide a formalized deep learning model for open set recognition.

\section*{Acknowledgment}

This work was supported by the National Research Foundation of Korea (NRF) grant funded by the Korean government (MSIT) (NRF-2019R1A2B5B01070358).

\bibliographystyle{unsrt}
\bibliography{reference}
\clearpage
\section{Supplementary Materials for "One-vs-Rest Network-based Deep Probability Model for Open Set Recognition"} \label{spplementary}

\subsection{The Structure of Proposed Model}

\begin{figure}[h]
  \centerline{\includegraphics[width=0.5\columnwidth]{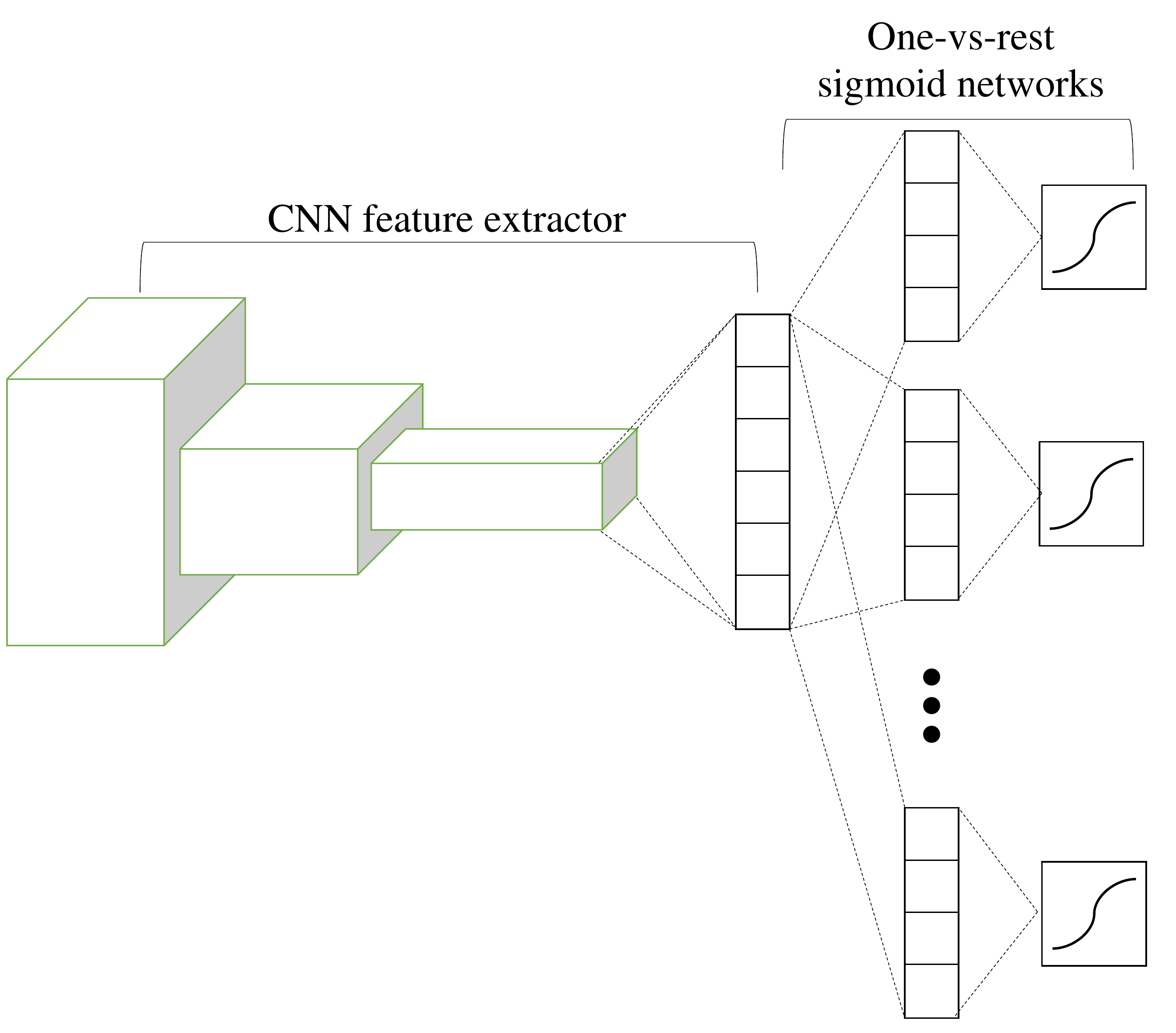}}
  \caption{The structure of proposed model for open set recognition.}
  \label{structure}
\end{figure}

The proposed model consists of a CNN feature extractor and the following one-vs-rest sigmoid networks. The one-vs-rest sigmoid networks, which use rectified linear unit (ReLU) activation for hidden layers, consist of a simple feedforward neural network that produces a sigmoid decision score for each target class, as shown in Fig. \ref{structure}.

\subsection{Illustrative Example Comparing Probability Models}

\begin{figure}[h]
  \centerline{\includegraphics[width=0.75\columnwidth]{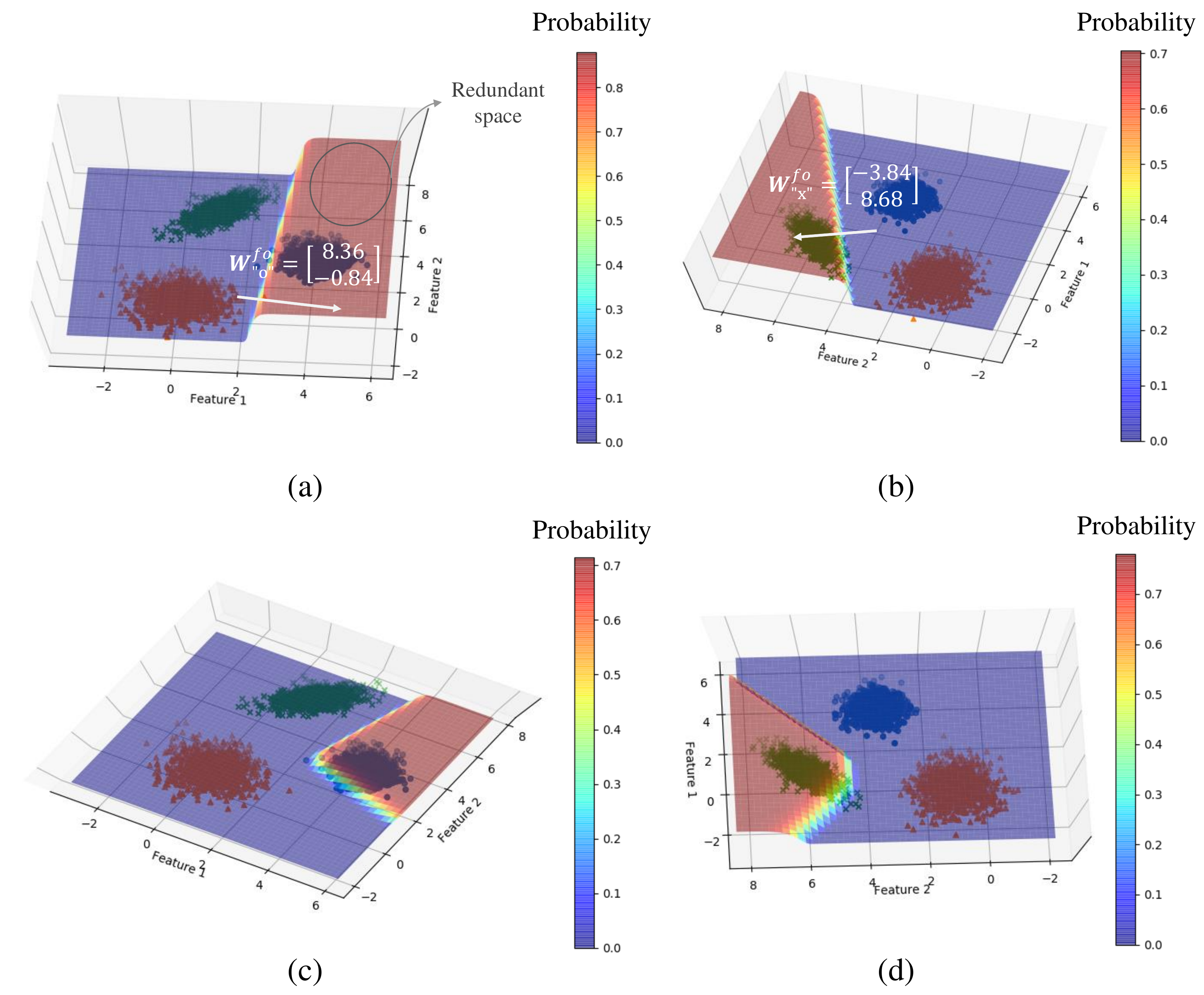}}
  \caption{The probability map estimated based on the single sigmoid output layer for (a) target class “o” and (b) target class “x”, and the probability map estimated based on the one-vs-rest sigmoid networks for (c) target class “o” and (d) target class “x”. For (a) and (b), a weight vector for the target class is included.}
  \label{illustrative}
\end{figure}

To observe the difference in probability mapping between single sigmoid output layer and one-vs-rest sigmoid networks, we designed an illustrative example. We assumed that there is the feature space, where three class examples are well separated linearly. The examples were used to train both mapping models, and the derived probability mappings are shown in Fig. \ref{illustrative}. Each one-vs-rest network had one hidden layer that consists of five ReLU hidden nodes and both models were trained to minimize cross-entropy loss, with final losses of $7\times10^{-4}$ and $9\times10^{-4}$ for the sigmoid layer and the one-vs-rest sigmoid networks. We applied the probability estimation technique proposed in Section \ref{model}.

\subsection{Datasets and Settings}

\subsubsection{EMNIST}

\textbf{KL divergence distribution comparison.} EMNIST consists of 47 handwritten character classes. To measure the KL divergence between known classes and unknown classes, we randomly selected 37 classes as known classes and the remaining 10 classes as unknown classes. For each known and unknown class, 2,400 data points originally provided for training in \cite{Cohen2017} were used. One-vs-rest networks with a single hidden layer containing ten nodes were used. The feature space is discretized into 10, 30, 50 and 70 areas.

\textbf{Open set classification performance comparison.} The proposed method and the four benchmarks were compared under various openness values. To vary the openness, we increased the number of known classes from 5 to 30 in intervals of 5 and randomly selected known classes. The remaining classes were assigned as unknown classes and were not used in training. The training and evaluation data that were originally provided for training and testing separately in \cite{Cohen2017} were used. We applied one-vs-rest networks with a single hidden layer containing ten nodes. We adopted cross-class validation, which uses only the training dataset, leaving five of the training classes as unknown classes, and using $20\%$ of the other classes of data only for evaluation to find the probability score limit $\theta$ and the ratio of extremes $\alpha$.

\subsubsection{LFW}

To evaluate the proposed and comparison methods on the LFW face dataset, we followed the data setting protocol introduced in \cite{Oza2019}. Specifically, a total of 12 classes containing more than 50 images were considered known classes, and each known class dataset was randomly divided into training and testing sets in an 80/20 ratio. The remaining 5,705 classes of the LFW dataset were considered unknown classes and were sorted in descending order. One-vs-rest networks with a single hidden layer containing ten nodes were used. Additionally, we introduced a data augmentation technique that includes shifting the image horizontally and vertically, rotating, and flipping into all the comparison methods. To find the optimal $\theta$ and $\alpha$, we executed cross-class validation, leaving four training classes containing the least data as unknown classes in validation and dividing the other eight training classes into training and testing splits of 80/20.

\subsection{Box-and-whisker plots for KL divergences}

\begin{figure}[h]
\centerline{
\begin{minipage}[b]{0.5\textwidth}
\begin{subfigure}[b]{0.49\textwidth}
	\includegraphics[width=\textwidth]{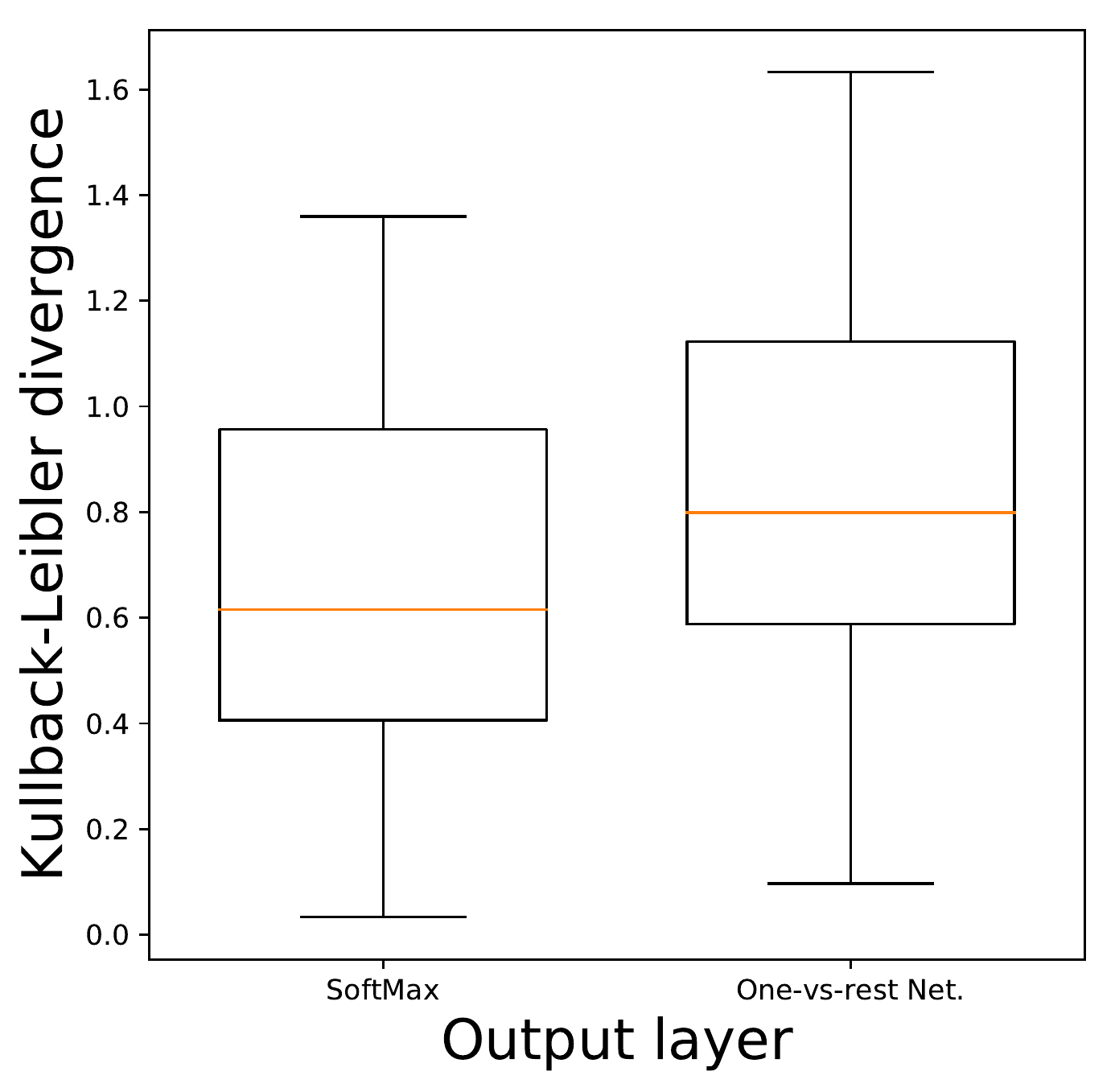}
\subcaption{4 areas}
\end{subfigure}
\begin{subfigure}[b]{0.49\textwidth}
\includegraphics[width=\textwidth]{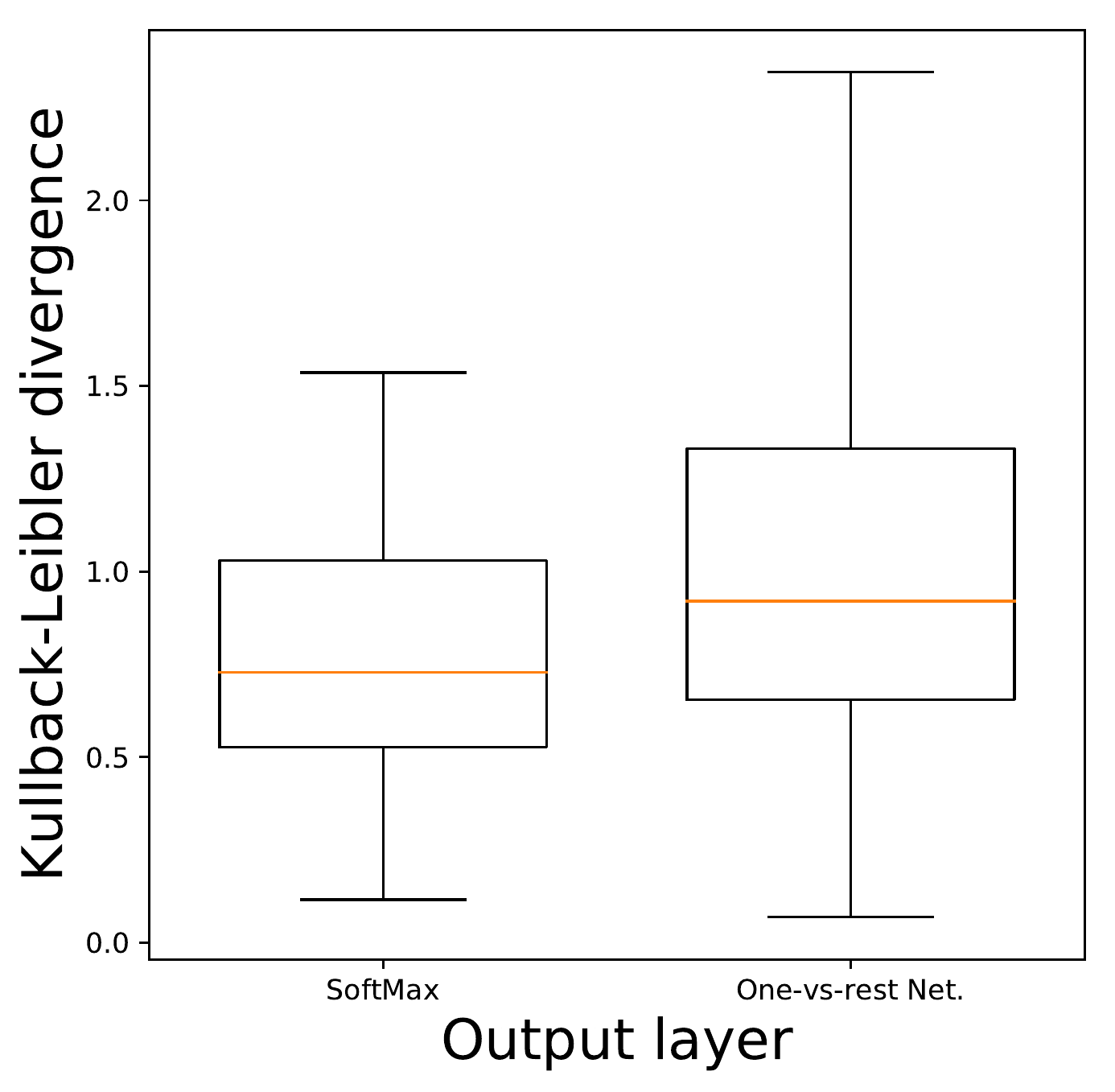}
\subcaption{6 areas}
\end{subfigure}
\end{minipage}
}
\centerline{
\begin{minipage}[b]{0.5\textwidth}
\begin{subfigure}[b]{0.49\textwidth}
	\includegraphics[width=\textwidth]{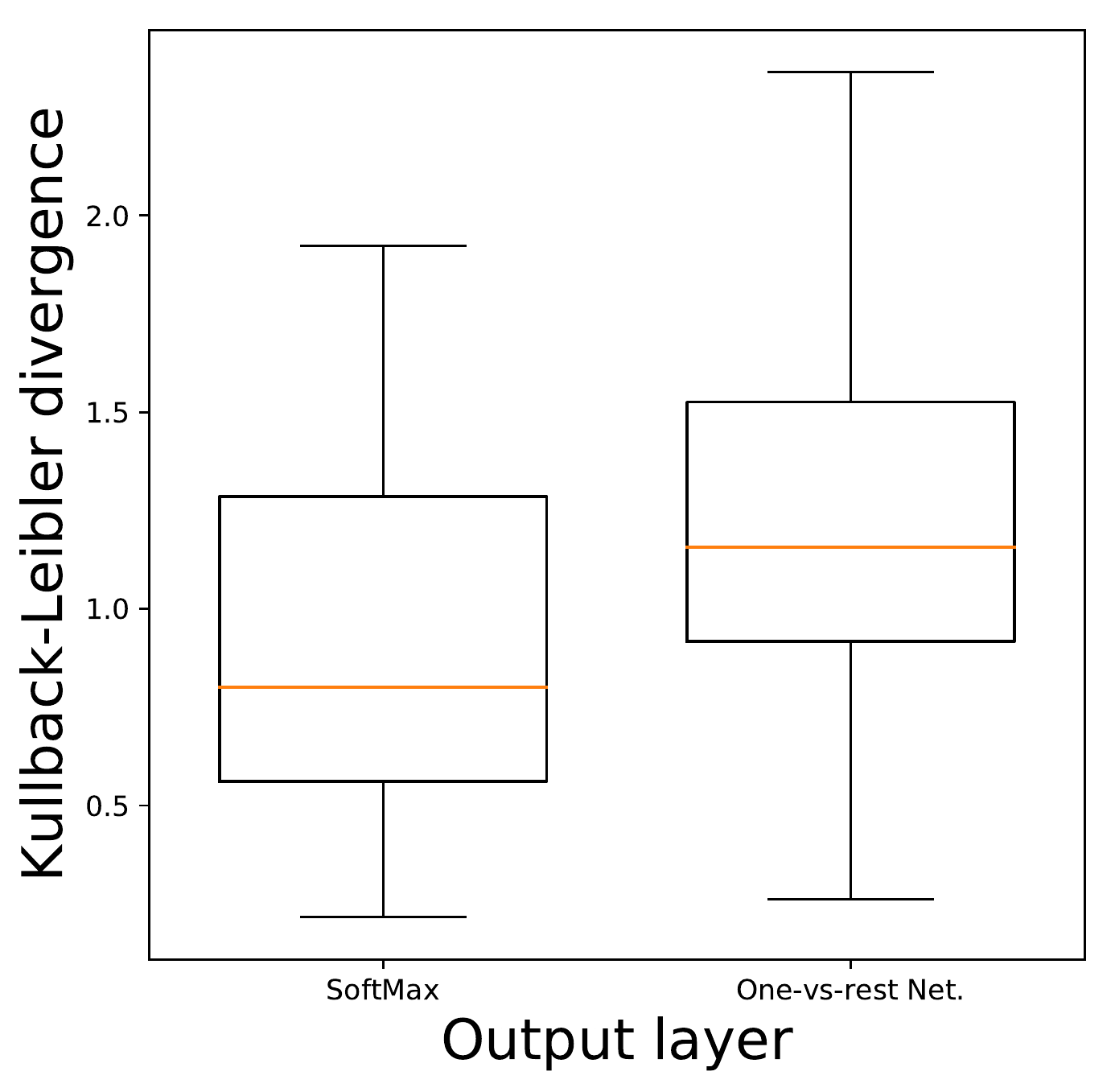}
\subcaption{8 areas}
\end{subfigure}
\begin{subfigure}[b]{0.49\textwidth}
\includegraphics[width=\textwidth]{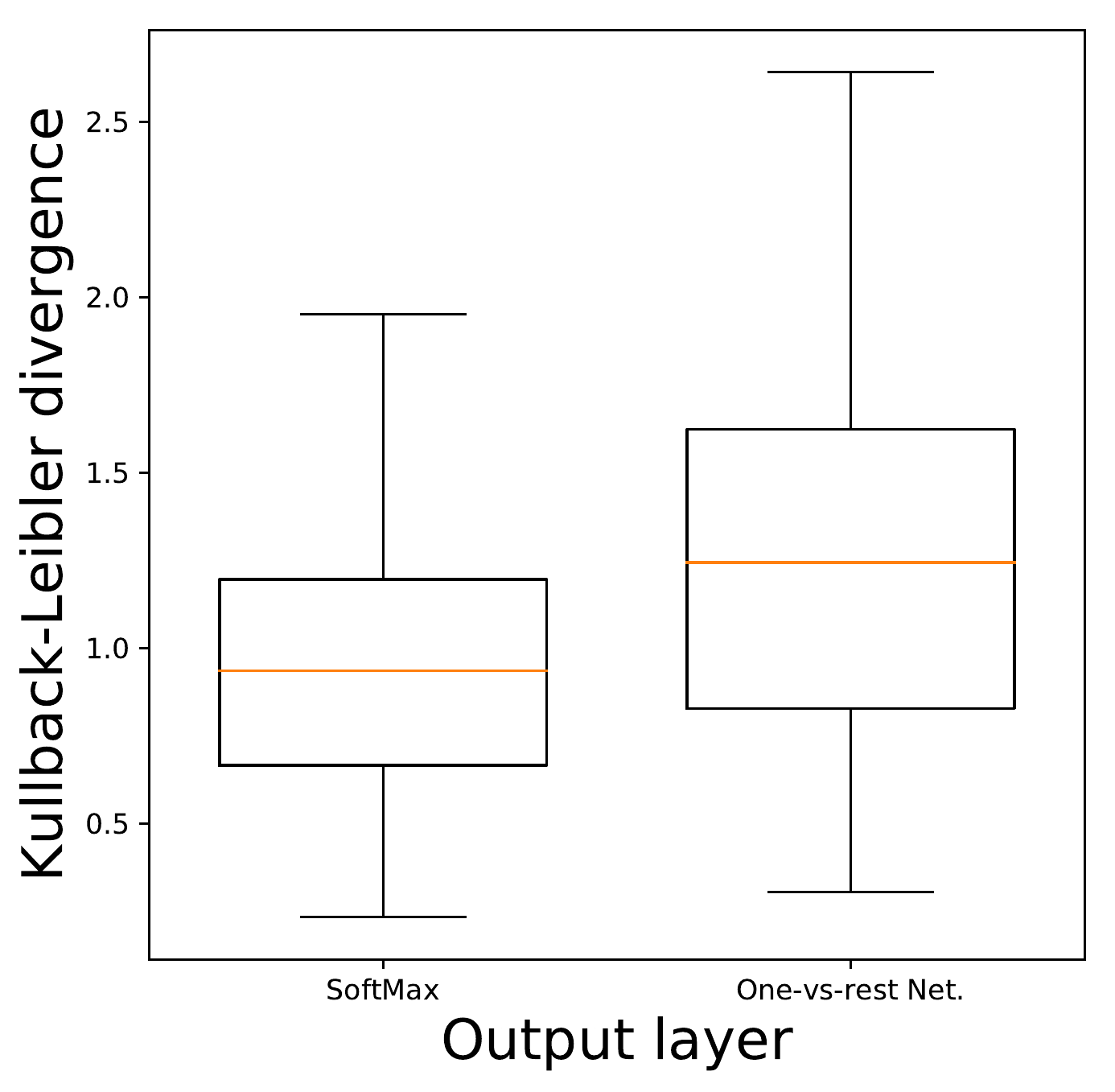}
\subcaption{10 areas}
\end{subfigure}
\end{minipage}
}
	\caption{Box-and-whisker plot showing the distributions of the KL divergences for the LFW dataset.}
  \label{LFW_box_and_whisker}
\end{figure}

\begin{figure}[h]
\centerline{
\begin{minipage}[b]{0.75\textwidth}
\begin{subfigure}[b]{0.32\textwidth}
	\includegraphics[width=\textwidth]{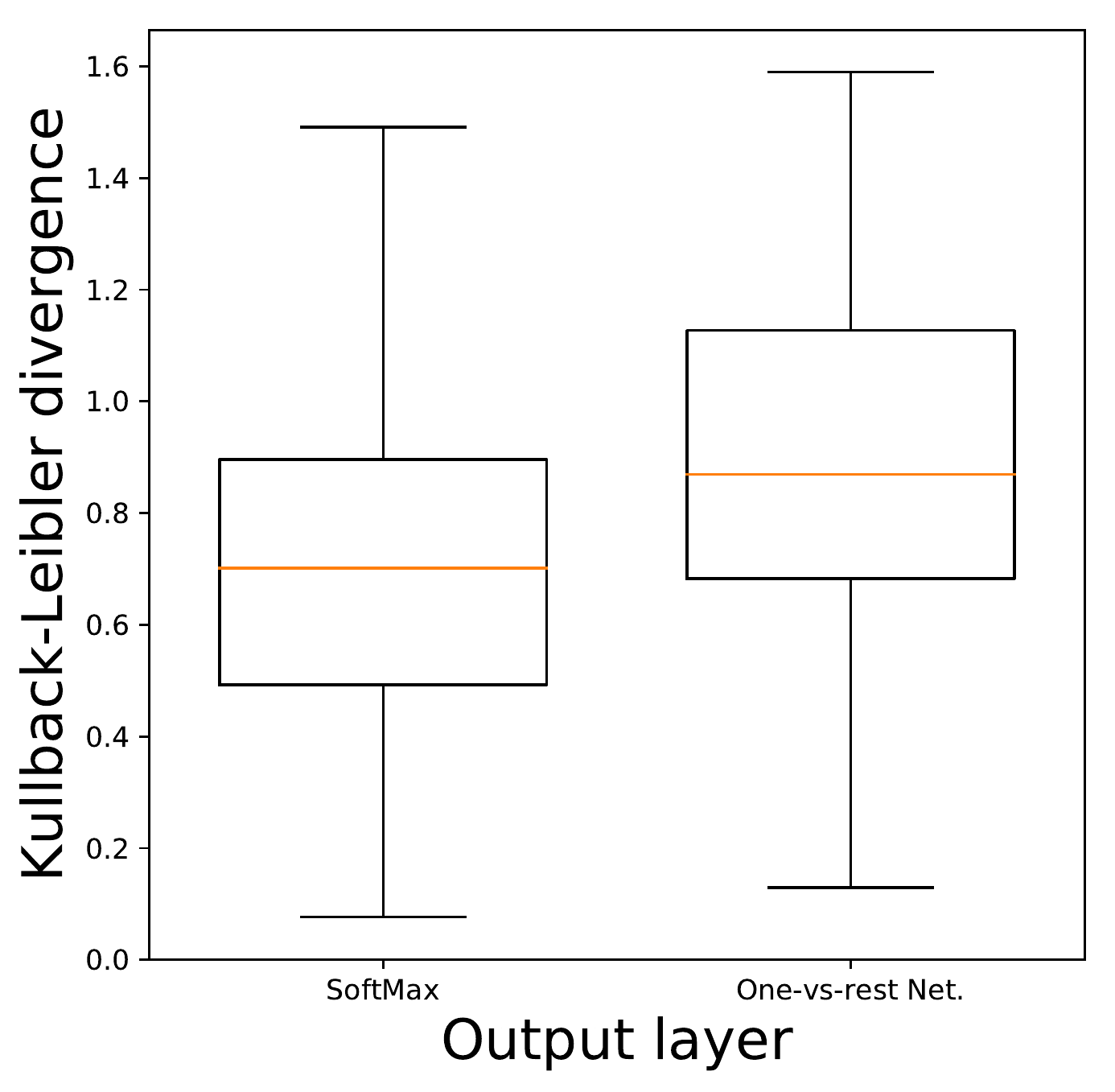}
\subcaption{5 areas}
\end{subfigure}
\begin{subfigure}[b]{0.32\textwidth}
\includegraphics[width=\textwidth]{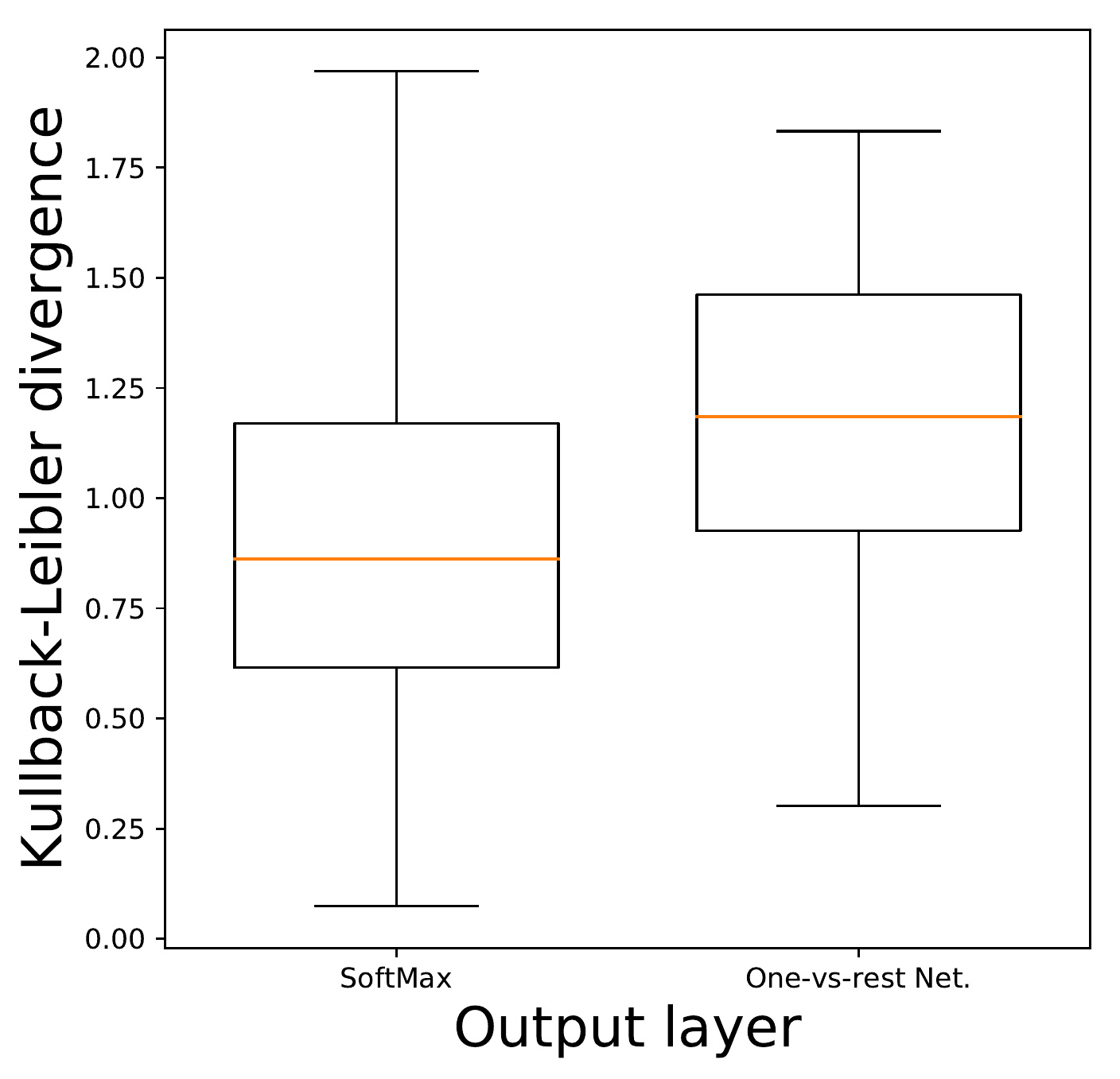}
\subcaption{10 areas}
\end{subfigure}
\begin{subfigure}[b]{0.32\textwidth}
	\includegraphics[width=\textwidth]{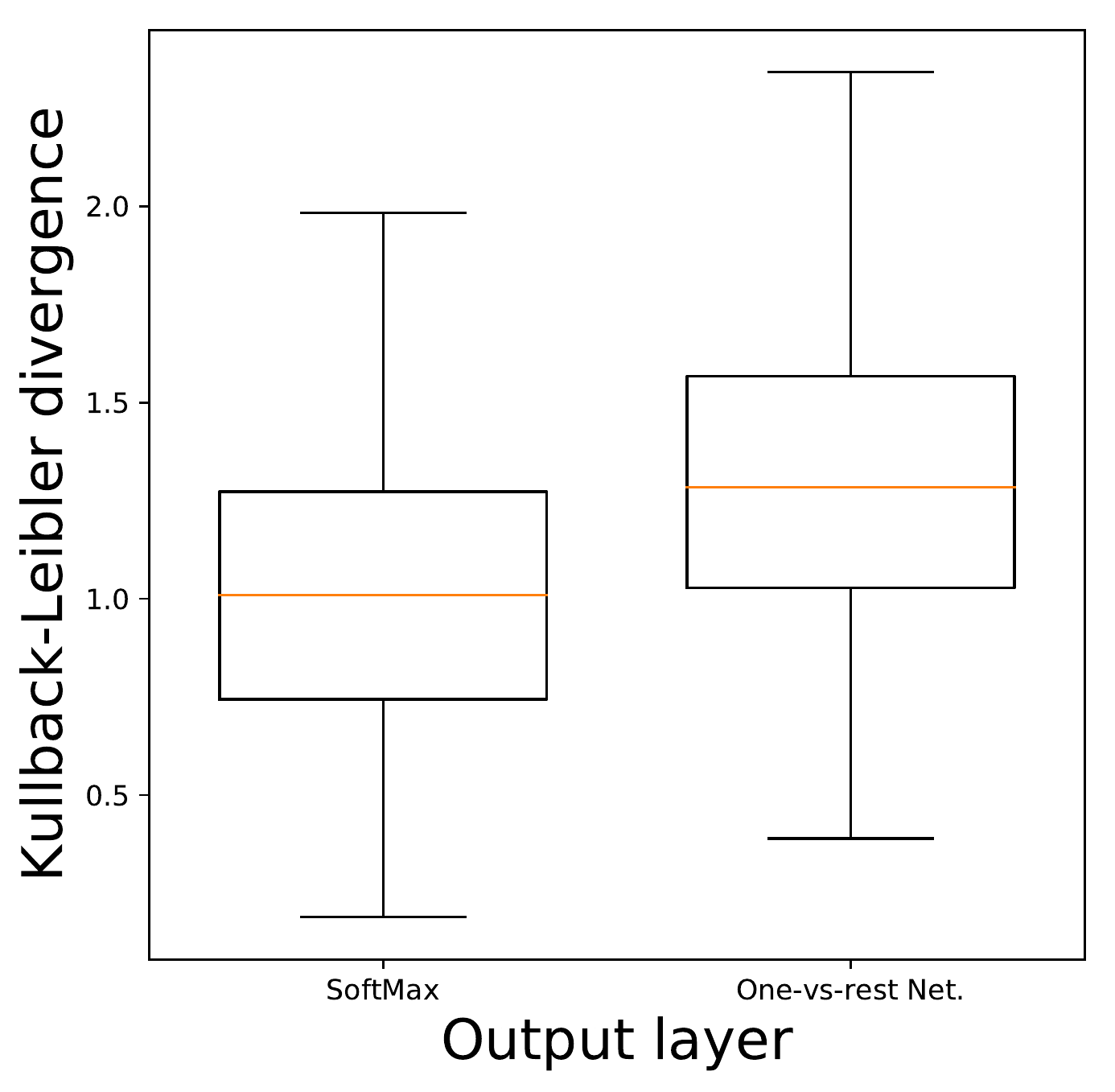}
\subcaption{15 areas}
\end{subfigure}
\end{minipage}
}
\centerline{
\begin{minipage}[b]{0.75\textwidth}
\begin{subfigure}[b]{0.32\textwidth}
	\includegraphics[width=\textwidth]{Cifar_KL_comparison_15.pdf}
\subcaption{15 areas}
\end{subfigure}
\begin{subfigure}[b]{0.32\textwidth}
\includegraphics[width=\textwidth]{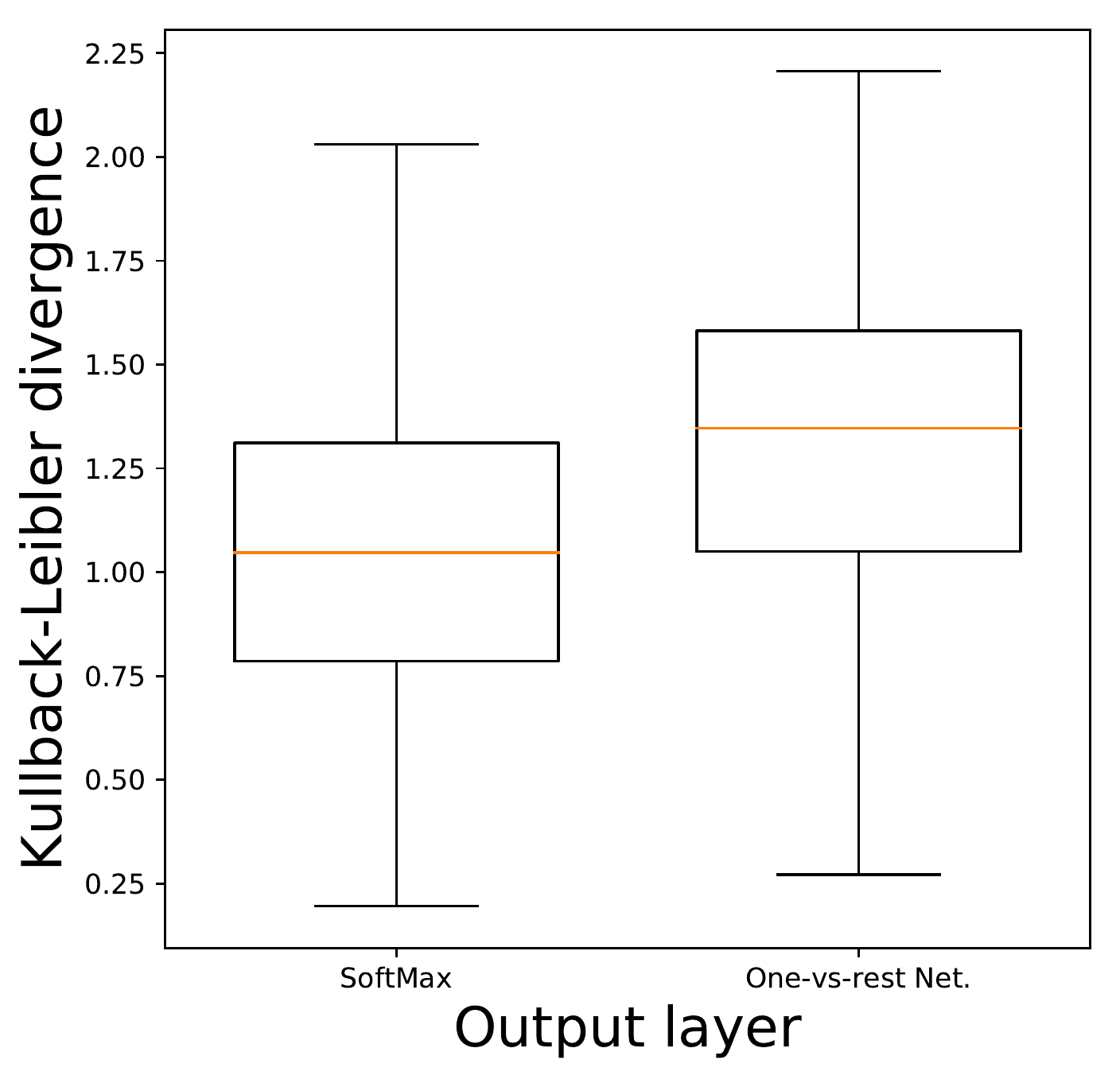}
\subcaption{20 areas}
\end{subfigure}
\begin{subfigure}[b]{0.32\textwidth}
	\includegraphics[width=\textwidth]{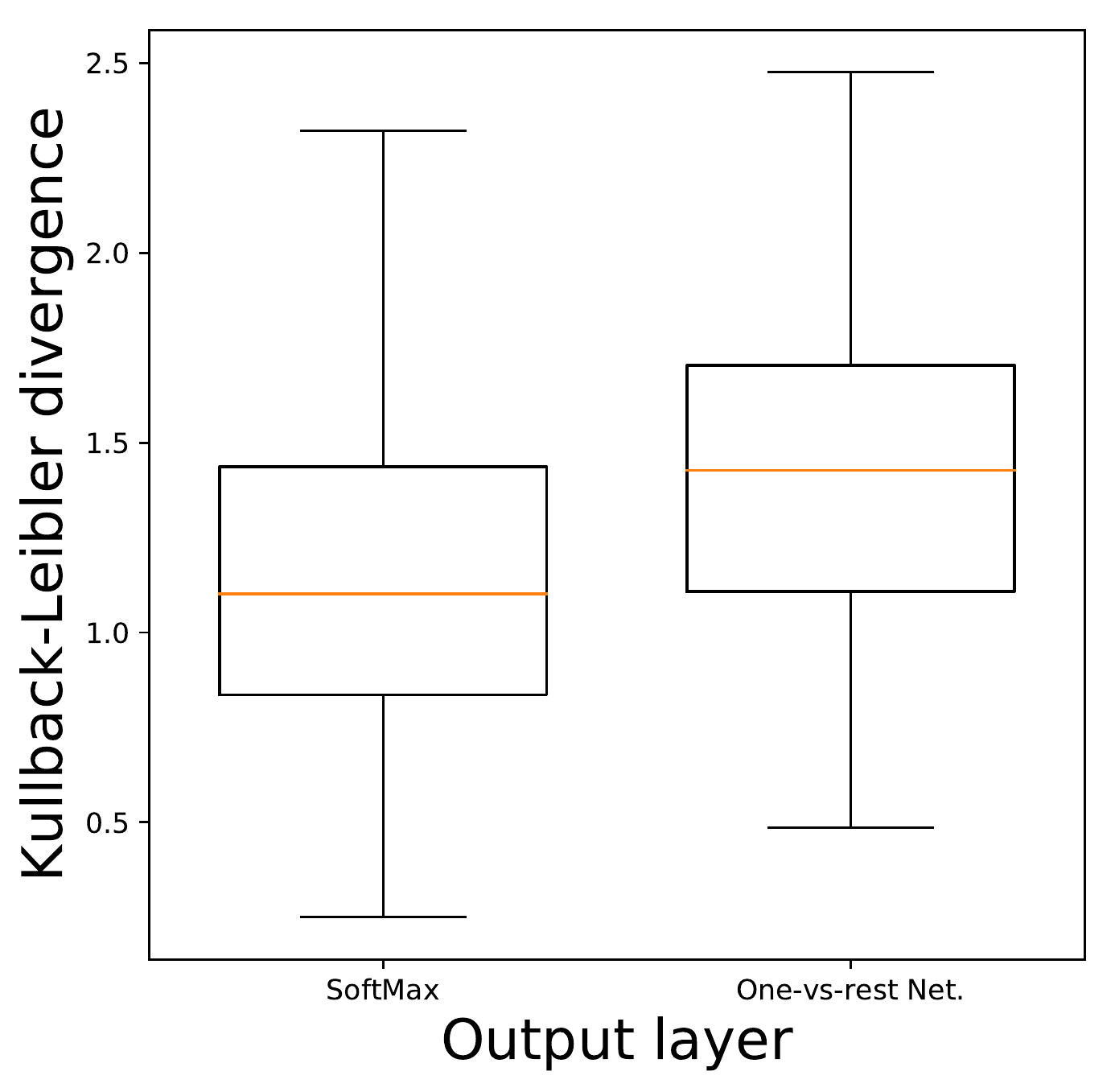}
\subcaption{30 areas}
\end{subfigure}
\end{minipage}
}
	\caption{Box-and-whisker plot showing the distributions of the KL divergences for the Cifar-100 dataset.}
  \label{Cifar_box_and_whisker}
\end{figure}

\end{document}